\documentclass[runningheads]{llncs}

\usepackage[T1]{fontenc}

\usepackage{graphicx}
\usepackage{xcolor}
\usepackage{booktabs}
\usepackage{amsmath}
\usepackage{amssymb}
\usepackage{subcaption}
  
\begin{document}

\title{Loss Landscape Topology Reveals Why Simple Baselines are Competitive at 3D Point Cloud Segmentation Under Class Imbalance\thanks{This work has received funding under Grant Agreement No. 101168067, GuardAI - Enhancing Robustness and Security of Edge AI Systems for Safety-Critical Applications, with support from the European Cybersecurity Competence Centre. The views and opinions expressed are, however, those of the author(s) only and do not necessarily reflect those of the European Union or the European Cybersecurity Competence Centre. Neither the European Union nor the European Cybersecurity Competence Centre can be held responsible for them. Computational resources were provided by the High Performance Computing facility of the University of Cyprus (UCY HPC).}}

\titlerunning{Loss Landscape Topology in 3D Segmentation}

\author{Antonis Savva\inst{1}\orcidID{0000-0002-0056-5680} \and Christos Kyrkou\inst{1}\orcidID{0000-0002-7926-7642} \and Theocharis Theocharides\inst{1,2}\orcidID{0000-0001-7222-9152}}

\authorrunning{A. Savva et al.}

\institute{
KIOS Research and Innovation Center of Excellence \\ 
\email{\{savva.d.antonis, kyrkou.christos\}@ucy.ac.cy} \and
Department of Electrical and Computer Engineering, \\
\email{theocharides.theocharis@ucy.ac.cy} \\
University of Cyprus, 1 Panepistimiou Avenue, 2109 Aglantzia, Nicosia
}

\maketitle

\begin{abstract}
Semantic segmentation of 3D point clouds faces severe class imbalance, yet the effectiveness of specialized imbalance-aware methods from 2D computer vision remains unclear in 3D contexts. We systematically evaluate 11 imbalance mitigation approaches across datasets with extreme (641:1) and moderate (56:1) imbalance ratios, revealing a surprising finding: standard cross-entropy with uniform weighting achieves competitive performance, typically within 0.8-3.3\% mIoU of specialized methods across architectures and datasets. Through multifaceted mechanistic analysis of error patterns, decision boundaries, and the geometry of the optimization landscape, our analyses suggest that imbalance severity shapes the topology, creating narrow solution basins under extreme imbalance and flat plateaus under moderate imbalance. This appears to constrain the effectiveness of loss-level modifications, as all methods must navigate these geometric constraints. Our findings offer practical guidance; standard cross-entropy provides a robust baseline, with specialized methods offering modest improvements (0.8-3.3\% mIoU) that vary by architecture and dataset but risk substantial degradation if poorly tuned. This work provides the first mechanistic explanation for why techniques proven effective in 2D do not readily transfer to point-based 3D point cloud segmentation, validated across two representative architectures.
\end{abstract}

\section{Introduction}\label{sec:intro}
Semantic segmentation of 3D point clouds from LiDAR data is essential for autonomous navigation, robotics \cite{He2025}, and environmental monitoring \cite{Bodoque2023}. As in most real-world datasets, LiDAR point clouds exhibit severe class imbalance, with the majority classes containing a large number of points, while the critical minority classes contain far fewer. This long-tailed distribution degrades the performance of rare classes, and accurate segmentation of these classes is essential for building reliable systems. In this context, imbalance mitigation strategies used in 2D computer vision are a natural way to adopt in 3D for improving performance, such as class-balanced loss based on the effective number of samples \cite{Cui2019}, focal loss \cite{Tsung-Yi2017}, label-distribution-aware margin \cite{Cao2019}, logit adjustment \cite{Menon2021}, seesaw loss \cite{Wang2021}, and balanced meta-softmax loss \cite{Ren2020}. These have demonstrated substantial improvements on benchmarks including CIFAR-LT, ImageNet-LT, iNaturalist for classification, COCO for object detection, and LVIS for instance segmentation.

Despite their success in 2D, whether these techniques transfer effectively to 3D remains underexplored. Unlike 2D images, which are distinguished by appearance, 3D point clouds encode semantics through geometry. Point-based architectures explicitly exploit this structure, potentially creating optimization landscapes where simple loss formulations suffice. This work examines whether specialized imbalance methods, when combined with deep learning architectures, offer similar benefits in 3D as they do in 2D and investigates the mechanisms underlying their performance.

Specifically, this work makes four primary contributions:

\begin{itemize}
    \item Systematic evaluation of 11 imbalance mitigation approaches (six loss re-weighting and five loss functions) across two diverse datasets. These are combined with confusion matrix analysis, decision boundary variability assessment, and loss landscape characterization to provide the first mechanistic explanation for why standard cross-entropy (CE) remains competitive in point-based 3D segmentation.
    \item Through detailed confusion matrix analysis, we show that specialized methods successfully increase minority class recall, but degrade precision, explaining why aggregate performance remains neutral despite targeting imbalance. For instance, aggressive reweighting reduces minority-to-majority errors by 5-10×, but inflates majority-to-minority errors by 10×.
    \item We measure how decision boundaries differ between CE and specialized methods, uncovering dataset-dependent landscape topology. On DALES, methods with low boundary variability from CE achieve 80-81\% mIoU while large deviations cause significant performance degradation (dropping to 68\%). On S3DIS, all methods cluster within 62.9-64.8\% mIoU regardless of boundary configuration, suggesting a flat landscape where method choice has minimal impact.
    \item We characterize local curvature via weight perturbation sensitivity and Hessian eigenvalue spectra, and find that all methods exhibit similar profiles within each dataset.
\end{itemize}

We analyze DALES \cite{Varney2020} (extreme imbalance, 641:1, outdoor aerial LiDAR) and S3DIS \cite{Armeni2016} (moderate imbalance, 56:1, indoor LiDAR) and reveal that CE with uniform weighting is competitive to specialized methods, with differences typically within 0.6-3.3\% mIoU. Our mechanistic analyses show that imbalance severity shapes the topology of the optimization landscape, with extreme imbalance creating narrow basins where CE's solution is favored and deviations are risky (in terms of performance degradation). In contrast, moderate imbalance produces flat landscapes where method choice has a lesser impact. These findings suggest that practitioners can rely on standard CE for point-based 3D segmentation tasks, though specialized methods may provide modest gains depending on the architecture. We validate these patterns across two point-based architectures (KPConv and RandLA-Net). Future work should extend this analysis to additional architectures (e.g., voxel- and transformer-based) and investigate combined data-level and loss-level approaches.

\section{Related Work}\label{sec:related_work}
\subsection{Point Cloud Semantic Segmentation}
Deep learning approaches for 3D point cloud segmentation can be categorized into projection-, voxel-, and point-based methods. Projection-based methods map 3D points onto multi-view or spherical images, enabling the use of standard CNNs but suffering from information loss and occlusions. Voxel-based methods partition space into regular grids for 3D convolutions, capturing local structure at the cost of high memory consumption \cite{He2025}. Point-based methods process raw point clouds directly, with PointNet proposing permutation-invariant feature learning \cite{Charles2017PointNet}, and PointNet++ introducing hierarchical feature aggregation \cite{Charles2017PointNetPP}. Subsequent advances include methods that: aggregate local features using multi-layer perceptrons to mitigate the drawbacks of random-point sampling methods \cite{Hu2020RandLANet}, learn convolutional kernels specifically for point clouds \cite{Wu2019PointConv}, and employ fully convolutional networks with kernel-point-based representations \cite{Thomas2019}.

\subsection{Class Imbalance}

Class imbalance mitigation methods include resampling strategies, cost-sensitive learning, and specialized loss functions such as focal loss (FL) \cite{Tsung-Yi2017}, label distribution aware margin loss (LDAM) \cite{Cao2019}, and logit adjustment (LADJ) \cite{Menon2021}, which have shown improvements on long-tailed benchmarks (iNaturalist, CIFAR-LT, ImageNet LT, LVIS) \cite{Gupta2019LVIS,Liu2019ImageNetLT,VanHorn2018iNaturalist,Zhang2023}.

Imbalance in 3D point clouds is intrinsic to the sensing modality, where LiDAR datasets exhibit extreme imbalance (e.g., Semantic-KITTI \cite{Behley2019}: road, sidewalk, parking comprise $\approx$40\% of points; signs, people, bicyclists <0.1\%) due to: (1) unequal physical extent (ground vs. small objects), (2) distance-dependent point density (distant objects are undersampled), and (3) environmental frequency (vegetation is ubiquitous; traffic signs are sparse). Pan \textit{et al}. \cite{Pan2023} conducted the first systematic analysis of imbalance characteristics in SemanticKITTI, categorizing classes by difficulty and showing that challenges arise not only from frequency but also from intra-class variation and inter-class geometric properties. However, their study did not systematically evaluate modern loss-level mitigation strategies and therefore, when and why these techniques succeed or fail. 

\subsection{Loss Landscape Analysis and Decision Boundaries}
To understand why imbalance-aware methods do not outperform standard CE in point-based 3D segmentation (despite their success in 2D), we need analytical tools to characterize how different loss-function formulations shape the optimization process and learned representations. Evaluation through aggregate metrics (e.g., mIoU) reveals \textit{what} performs well but not \textit{why}, leaving the mechanisms underlying method effectiveness unexplained.

Understanding optimization landscapes has become essential in deep learning theory for precisely this purpose. Loss landscape visualization and Hessian eigenvalue analysis characterize local curvature, with flatness linked to better generalization \cite{Li2018}, enabling the assessment of how different choices (e.g., batch size, skip connections) guide optimization. Complementarily, decision boundary analysis quantifies how networks partition the input space (Lei \textit{et al}. introduced a variability metric measuring boundary stability across training runs \cite{Lei2025}), allowing us to determine whether specialized methods alter learned decision boundaries compared to CE. Together, these approaches provide a mechanistic lens to examine whether performance differences (or lack thereof) stem from landscape geometry, different geometric partitions, or both. In this study, we employ both frameworks to provide the first comprehensive mechanistic analysis of class imbalance mitigation in point-based 3D point cloud segmentation.

\section{Methodology}\label{sec:methodology}
Our work addresses literature gaps by: (1) evaluating 11 methods across datasets with contrasting imbalance severities; (2) conducting mechanistic analysis through decision boundary variability; (3) performing confusion matrix analysis that identifies precision-recall trade-offs underlying specialized methods' failure to improve aggregate metrics; and (4) characterizing loss landscape geometry via network's weights perturbation and Hessian eigenvalues. To the best of our knowledge, this is the first work to connect the topology of the optimization landscape to the effectiveness of imbalance mitigation in point-based 3D segmentation.

\subsection{Problem Setup}
Given a point cloud $\mathcal{P} = \{p_i\}_{i=1}^N$ with $p_i \in \mathbb{R}^{3+d}$ (coordinates plus optional features), the task is to assign each point a semantic label $y_c \in \{1,\ldots,C\}$ with $C$ being the number of classes and $N$ the number of points in the dataset.

\subsection{Evaluated Methods}
\subsubsection{Reweighting schemes.}
We modify the standard CE loss as \\ $\mathcal{L}_{\text{CE}} = -\sum_{i=1}^{N} w_{c} \log(p_{y_i})$, where $p_{y_i}$ is the predicted probability for the actual class and $w_c$ is the class weight. With $n_c$ denoting the number of training points in class $c$, the six schemes are: (1)~\textit{uniform (uni)} $w_c = 1$ (baseline), (2)~\textit{inverse frequency (invf)} $w_c = N/n_c$, (3)~\textit{class-balanced (cb)} loss based on the effective number of samples $w_c = (1-\beta)/(1-\beta^{n_c})$ with $\beta=0.9$ \cite{Cui2019,Pan2023}, (4)~\textit{inverse logarithm (invl)} $w_c = 1/\log(n_c)$, (5)~\textit{inverse power (invp)} $w_c = 1/n_c^{\gamma}$ with $\gamma=0.1$, and (6)~\textit{complementary frequency (comf)} $w_c = 1 - n_c/N$ \cite{Prakash2023}. Weights are normalized to sum to 1 for all schemes except the uniform scheme.

\subsubsection{Loss functions.}
We evaluate: (1)~\textit{FL}: $\mathcal{L}_{\text{FL}} = -\sum_i (1-p_{y_i})^\gamma \log(p_{y_i})$ with $\gamma=1$, which down-weights well-classified examples \cite{Tsung-Yi2017}; (2)~\textit{LDAM}, which enforces class-dependent decision margins $\Delta_c \propto 1/n_c^{1/4}$ \cite{Cao2019}; (3)~\textit{LADJ}, which adjusts logits as $\tilde{z}_c = z_c - \tau \log \pi_c$ with class prior $\pi_c = n_c/N$ and $\tau=0.3$ \cite{Menon2021}; (4)~\textit{BalSoft}, which incorporates class frequencies into the softmax calculation \cite{Ren2020}; and (5)~\textit{Seesaw Loss}, which dynamically balances between mitigation and compensation factors, with the former reducing the penalty for the tail classes and the latter increasing the penalty when misclassifications occur \cite{Wang2021}.

\subsection{Architecture and Datasets}
We employ Kernel Point Convolution (KPConv) \cite{Thomas2019}, that uses rigid and deformable kernels for geometric convolutions to segment DALES (outdoor aerial LiDAR with eight semantic classes) \cite{Varney2020} and S3DIS (indoor scans with 13 semantic classes) \cite{Armeni2016}, as well as RandLA-Net that employs random sampling coupled with a local feature aggregation module ~\cite{Hu2020RandLANet}. More information on the specific architectures used, the training protocol, and the datasets is provided in Sections S1 and S2 of the supplementary.

\subsection{Analysis Methods}
While prior work has characterized the multi-faceted nature of imbalance in 3D segmentation \cite{Pan2023}, a systematic evaluation of multiple mitigation strategies with mechanistic analysis remains lacking. To address this gap, we employ three complementary techniques: (1)~confusion matrix analysis, which reveals precision-recall trade-offs driving performance differences; (2)~decision boundary variability, which quantifies how specialized losses alter learned partitions of the input space compared to CE; and (3)~loss landscape topology characterization via Hessian eigenvalues and weight perturbation, which assesses whether methods converge to qualitatively different regions (sharp vs. flat minima). Together, these analyses provide mechanistic insight into why standard CE remains competitive in point-based 3D segmentation.

\subsubsection{Confusion matrix.} For each method, we calculate the full confusion matrix on the test set and analyze: (1)~precision and recall for each class; and (2)~the percentage of points from majority classes that are assigned to minority classes and vice versa.

\subsubsection{Decision boundary variability.} To quantify how different methods alter the learned decision boundary (DB) compared to CE, we adopt the decision boundary analysis framework from \cite{Lei2025}. For a trained model $f_{\theta}$, we define the predicted label assignment as $y_c = \arg\max_c f_{\theta}(\mathbf{x})_c$. The decision boundary variability between two methods (method $A$ with parameters $\theta_A$ and uniform weighting with parameters $\theta_{\text{uni}}$) measures the proportion of test points receiving different label predictions:
\begin{equation}
    \text{DB-var}(A, \text{uni}) = \frac{1}{N} \sum_{i=1}^{N} \mathbb{I}\left[\arg\max_c f_{\theta_A}(\mathbf{x}_i)_c \neq \arg\max_c f_{\theta_{\text{uni}}}(\mathbf{x}_i)_c\right], \label{eq:db_variability}
\end{equation}
where $\mathbb{I}[\cdot]$ is the indicator function. Low variability indicates decision boundaries similar to those of CE, while high variability suggests substantial boundary alterations.

\subsubsection{Loss landscape topology.}\label{sec:ll_topology} We characterize the local curvature properties of the optimization landscape at converged model parameters $\theta^*$ by measuring sensitivity to weight perturbations. This analysis reveals whether a solution occupies a sharp, narrow minimum or a broad, flat basin; properties linked to generalization \cite{Li2018}.

Our approach samples $K$ random perturbation vectors from the tangent space at $\theta^*$. Each vector $v_k \sim \mathcal{N}(0, I)$ is generated independently and then scaled using filter-normalized magnitude matching \cite{Li2018}. Specifically, for each weight tensor $W_j$ in the network, its corresponding perturbation component $v_{k,j}$ is rescaled as:
\begin{equation}
    \hat{v}_{k,j} = v_{k,j} \cdot \frac{\|W_j\|_F}{\|v_{k,j}\|_F},
\end{equation}
where $\|\cdot\|_F$ denotes the Frobenius norm. This normalization ensures perturbations are proportional to each layer's natural scale, preventing disproportionate impacts from layers with different weight magnitudes.

We then evaluate the training loss at perturbed parameters $\tilde{\theta}_k = \theta^* + \rho \hat{v}_k$ for different values $\rho$. The loss deviation quantifies the landscape flatness:
\begin{equation}
    \delta_k = \mathcal{L}_{\text{train}}(\tilde{\theta}_k) - \mathcal{L}_{\text{train}}(\theta^*). \label{eq:loss_difference}
\end{equation}

Lower values $\delta_k$ characterize a flatter minimum, indicating that the model is less sensitive to small (weight) perturbations. We use $K=20$ random directions to provide a good balance between computational cost and statistical reliability.

To characterize local curvature, we also compute the top-10 eigenvalues of the Hessian matrix $\mathbf{H} = \nabla^2_{\theta} \mathcal{L}(\theta^*)$ at converged solutions $\theta^*$. We use the Lanczos algorithm, which requires only Hessian-vector products computed directly via automatic differentiation \cite{Li2018}. 

\section{Results}\label{sec:results}

\subsection{Evaluation Metrics}
Evaluation results are reported for a single training run with test-time predictions aggregated using 10 voting passes over the test datasets. Performance is measured using per-class IoU and mean IoU (mIoU):
\begin{equation}
    IoU_i = \frac{cm_{ii}}{cm_{ii} + \sum_{j \neq i} cm_{ij} + \sum_{k \neq i} cm_{ki}}, \quad
    mIoU = \frac{1}{C} \sum_{i=1}^{C} IoU_i,
    \label{eq:metrics}
\end{equation}
where $cm_{ij}$ denotes the elements of the confusion matrix.

\subsection{Overall performance}\label{sec:overall_perf}
\begin{table}[b]
\begin{center}
\vspace{-0.5cm}
\caption{Per-class performance on the DALES dataset.}\label{tab:full_iou_dales}
    \resizebox{\textwidth}{!}{%
        \begin{tabular}{|l||c|c|c|c|c|c|c|c|c|}

        \hline
        Method & mean & \textit{ground} & \textit{vegetation} & \textit{cars} & \textit{trucks} & \textit{power lines} & \textit{fences} & \textit{poles} & \textit{buildings} \\ 

        \hline
        \hline
        uni     & 80.047 & 96.507 & 93.781 & 85.143 & 42.650 & 94.029 & 61.092 & 72.262 & 94.915 \\
        
        \hline
        invf    & 67.780 & 95.885 & 91.111 & 71.534 & 32.470 & 86.406 & 31.041 & 40.163 & 93.632 \\
        cb      & 78.534 & 96.445 & 93.474 & 85.103 & 43.480 & 94.763 & 57.071 & 63.121 & 94.816 \\
        invl    & 80.692 & 96.488 & 93.846 & 84.828 & 43.504 & 94.476 & 62.659 & 74.936 & 94.802 \\
        invp    & 80.861 & 96.518 & 93.795 & 85.137 & 44.374 & 94.647 & 63.169 & 74.299 & 94.945 \\
        comf    & 80.715 & 96.474 & 93.759 & 85.192 & 43.103 & 94.646 & 63.280 & 74.335 & 94.932 \\

        \hline
        FL	    & 80.039 & 96.495 & 93.767 & 85.025 & 43.339 & 93.790 & 62.402 & 70.609 & 94.884 \\
        LDAM	& 80.629 & 96.510 & 93.828 & 85.260 & 44.550 & 94.510 & 62.524 & 72.894 & 94.957 \\
        LADJ	& 79.598 & 96.561 & 93.665 & 84.088 & 42.359 & 94.277 & 59.860 & 70.928 & 95.047 \\
        BalSoft	& 68.136 & 96.126 & 91.771 & 74.583 & 22.637 & 93.752 & 33.658 & 38.856 & 93.708 \\
        Seesaw	& 80.360 & 96.538 & 93.826 & 85.091 & 43.641 & 93.990 & 62.984 & 71.885 & 94.924 \\
        
        \hline \hline
        Range & 13.081 & 0.676 & 2.735 & 13.726 & 21.913 & 8.357 & 32.239 & 36.080 & 1.415 \\
        \hline
        \end{tabular}%
    }
\end{center}
\vspace{-0.6cm}
\end{table}

Tables~\ref{tab:full_iou_dales} and~\ref{tab:full_iou_S3DIS} present per-class and mean IoU for all evaluated methods on DALES (\textit{unknown} class is excluded when computing mIoU as per standard practice) and S3DIS, respectively. On DALES, uniform weighting (uni) achieves 80.05\% mIoU, competitive with the best-performing methods (invp: 80.86\%, invl: 80.69\%, comf: 80.72\%, LDAM: 80.63\%), while extreme reweighting (invf) and balanced-softmax loss (BalSoft) significantly reduce performance (67.78\%, 68.14\%). The differences between competitive methods span only 0.9\% mIoU. On S3DIS, all methods cluster within 62.92-64.85\% mIoU (1.93\% range), with uniform weighting (63.1\%) near the center; this narrow range across diverse loss formulations is itself a key finding, demonstrating that method choice has limited impact in this setting. The last row shows the performance range across methods, revealing that minority classes exhibit greater variance, while majority classes remain stable.

To ensure the observed patterns are not artifacts of random seed selection, we trained three additional runs with different seeds (i.e., seeds 1, 2, and 3) for uniform weighting (uni) and the best-performing method on each dataset (invp for DALES, BalSoft for S3DIS). The results are presented in Section S3 of the supplementary, confirming that method rankings remain stable across seeds; uniform weighting's competitive performance is not a statistical artifact, and that standard deviation increases for minority classes without altering the fundamental pattern of tight clustering among competitive methods.

To assess whether performance patterns generalize beyond KPConv, we repeated experiments using RandLA-Net \cite{Hu2020RandLANet}, which employs random sampling rather than KPConv's structured potential-based method. RandLA-Net exhibits identical patterns with slightly larger performance range, i.e., on DALES, uniform weighting achieves 76.76\%, while the best method (LDAM) achieves 79.17\% ($+$2.41\%), and on S3DIS, uniform achieves 61.38\% while LDAM achieves 64.70\% ($+$3.32\%). While mechanistic analyses (Sections \ref{sec:confusion_matrix}-\ref{sec:ll_topology_results}) focus on KPConv, the consistency of performance outcomes across architectures suggests the competitive performance of uniform CE is not architecture-specific. Complete results appear in Tables S2-S3 of the supplementary.

\begin{table}[t]
\begin{center}
\vspace{-0.5cm}
\caption{Per-class performance on the S3DIS dataset.}\label{tab:full_iou_S3DIS}

    \resizebox{\textwidth}{!}{%
        \begin{tabular}{|l||c|c|c|c|c|c|c|c|c|c|c|c|c|c|}

        \hline
        Method & mean & \textit{ceiling} & \textit{floor} & \textit{wall} & \textit{beam} & \textit{column} & \textit{window} & \textit{door} & \textit{chair} & \textit{table} & \textit{bookcase} & \textit{sofa} & \textit{board} & \textit{clutter} \\

        \hline
        \hline
        uni     & 63.097 & 93.693 & 98.514 & 80.862 & 0.000 & 21.341 & 43.934 & 61.597 & 87.347 & 79.061 & 71.098 & 65.612 & 59.560 & 57.642 \\

        \hline
        invf	& 63.683 & 92.264 & 98.314 & 80.297 & 0.000 & 25.512 & 46.554 & 60.456 & 87.337 & 79.534 & 71.026 & 71.432 & 60.282 & 54.871 \\
        cb	    & 63.556 & 92.649 & 98.392 & 80.499 & 0.000 & 24.290 & 45.111 & 62.371 & 87.402 & 78.244 & 70.891 & 67.834 & 62.327 & 56.221 \\
        invl	& 63.534 & 93.449 & 98.498 & 80.746 & 0.000 & 23.703 & 45.545 & 59.198 & 87.668 & 79.424 & 71.799 & 67.024 & 60.248 & 58.635 \\
        invp	& 62.915 & 92.995 & 98.416 & 80.498 & 0.000 & 23.474 & 46.459 & 60.386 & 86.864 & 78.853 & 70.282 & 60.821 & 61.380 & 57.465 \\
        comf	& 63.534 & 92.937 & 98.398 & 81.069 & 0.000 & 21.561 & 46.451 & 66.563 & 87.381 & 79.507 & 70.915 & 62.936 & 61.760 & 56.469 \\
        
        \hline
        FL	    & 63.247 & 92.213 & 98.394 & 79.701 & 0.000 & 22.118 & 44.992 & 60.418 & 87.755 & 79.353 & 70.239 & 67.789 & 62.686 & 56.555 \\
        LDAM	& 63.268 & 92.684 & 98.476 & 80.578 & 0.000 & 24.958 & 45.116 & 62.027 & 87.850 & 78.675 & 71.300 & 63.466 & 60.798 & 56.553 \\
        LADJ	& 63.999 & 92.834 & 98.405 & 81.447 & 0.000 & 24.748 & 49.772 & 62.241 & 87.156 & 78.881 & 71.915 & 64.341 & 63.757 & 56.492 \\
        BalSoft	& 64.848 & 93.138 & 98.383 & 82.602 & 0.000 & 28.378 & 54.368 & 64.392 & 86.916 & 77.966 & 71.283 & 66.948 & 63.054 & 55.598 \\
        Seesaw	& 63.610 & 92.922 & 98.453 & 80.586 & 0.000 & 21.489 & 47.755 & 62.613 & 87.691 & 78.743 & 70.777 & 67.915 & 62.434 & 55.549 \\       

        \hline \hline
        Range & 1.933 & 1.480 & 0.200 & 2.901 & 0.000 & 7.037 & 10.434 & 7.365 & 0.986 & 1.568 & 1.676 & 10.611 & 4.197 & 3.764 \\
        \hline
        \end{tabular}%
    }
\end{center}
\vspace{-0.6cm}
\end{table}

\subsection{Per-Class Performance Analysis}

\begin{figure}[b!]
    \includegraphics[width=\textwidth]{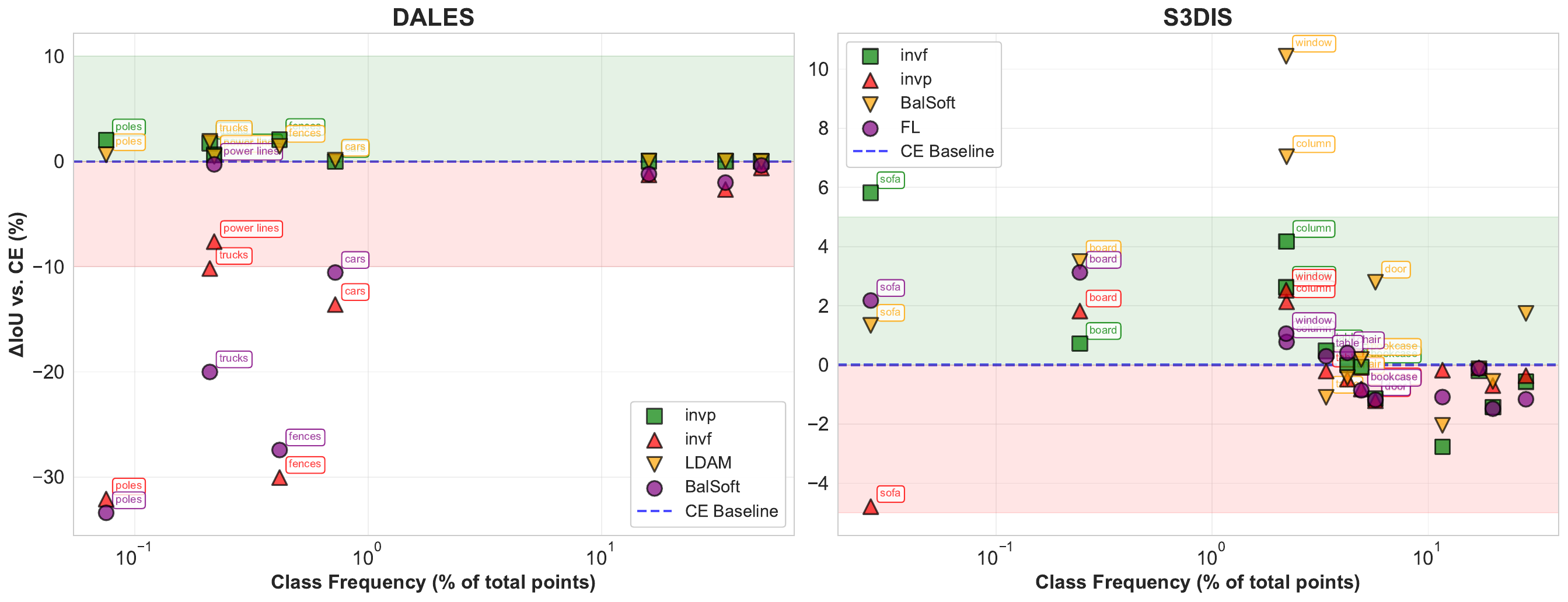}
    \caption{Class frequency vs. performance change relative to baseline (method - CE) for DALES (left) and S3DIS (right). Points represent individual classes for selected methods, with annotations showing class names where $|\Delta \text{IoU}| > 5\%$ (DALES) or $|\Delta \text{IoU}| > 3\%$ (S3DIS), or frequency $<6\%$. Shaded regions indicate improvement (green) and degradation (red) zones. DALES exhibits extreme method sensitivity (a significant reduction in rare classes with invf and BalSoft), whereas S3DIS shows tight clustering across methods.} 
    \label{fig:freq_vs_deltaiou}
    \vspace{-0.5cm}
\end{figure}

To understand how the different methods affect individual classes, we analyze the relationship between class frequency and performance gains relative to CE for KPConv. Figure~\ref{fig:freq_vs_deltaiou} presents class frequency against $\Delta$IoU (performance difference from CE) for representative methods (i.e., best and least performing methods among loss re-weighting and loss functions). Here, classes are annotated where changes exceed 5\% (DALES) or 3\% (S3DIS), or where frequency is below 6\%.

On DALES, extreme reweighting (invf) and BalSoft show significant degradation on minority classes, with \textit{poles} and \textit{fences} suffering losses exceeding $-30\%$ IoU. Conversely, mild reweighting (invp) and LDAM loss achieve modest gains (+2-3\%) on these same classes, while maintaining majority class performance (Table \ref{tab:full_iou_dales}). There is no consistent correlation between class frequency and benefit from the analyzed methods; some rare classes improve while others degrade substantially. 

On S3DIS, all methods exhibit much tighter clustering around the baseline, with $\Delta$IoU values predominantly within $\pm$5\%. The best-performing method (BalSoft) shows improvements on \textit{window} and \textit{column} (+10\%, +7\%); however, invp exhibits degradation on \textit{sofa} (-5\%). This inconsistency across classes, combined with the small magnitude of changes, reveals why mIoU differences remain within 2\% despite varied loss formulations.

\begin{figure}[t]
    \includegraphics[width=\textwidth]{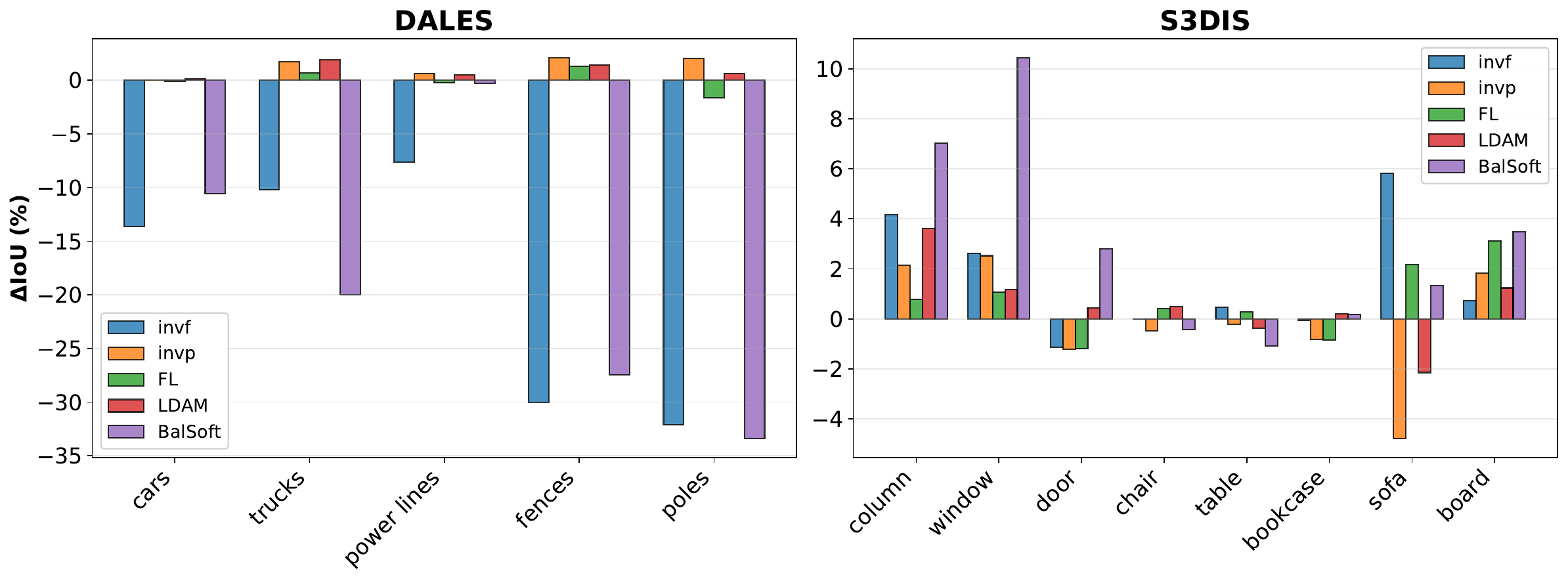}
    \caption{Performance change ($\Delta$IoU) for minority classes (frequency <6\%) on DALES (left) and S3DIS (right). For DALES, aggressive reweighting results in drastic degradation, whereas other methods yield modest gains. For S3DIS, all methods exhibit lower differences, indicating minimal impact of method choice.}
    \label{fig:minority_comparison}
    \vspace{-0.5cm}
\end{figure}

Figure~\ref{fig:minority_comparison} focuses specifically on minority classes (<6\% frequency). On DALES, aggressive reweighting drastically harms rare class performance, whereas on S3DIS, even substantial loss modifications yield only marginal and inconsistent improvements. These results suggest that specialized methods do not systematically benefit minority classes in point-based 3D segmentation, contrary to their established effectiveness in 2D image classification \cite{Cao2019,Cui2019,Tsung-Yi2017,Menon2021,Ren2020,Wang2021}.

\subsection{Confusion Matrix}\label{sec:confusion_matrix}
To understand why specialized methods do not improve aggregate performance despite targeting minority classes, we analyze error patterns through confusion matrices. Table \ref{tab:confusion_analysis} quantifies two error types, i.e., majority-to-minority and minority-to-majority misclassifications. In DALES, a fundamental trade-off with uniform weighting is revealed, achieving balanced error rates (0.15\% majority-to-minority, 12.78\% minority-to-majority). Aggressive reweighting (invf) and BalSoft dramatically reduce minority-to-majority errors (1.58\%, 1.89\%); thus, successfully increasing minority class recall, but inflate majority-to-minority errors by 10× (1.58\%, 1.50\%), degrading minority class precision. This precision collapse explains their low mIoU scores (67.78\%, 68.14\%). Conversely, competitive methods (invl, invp, comf, LDAM) maintain low majority-to-minority contamination ($\leq$0.19\%), preserving precision, while accepting slightly higher minority-to-majority errors (10.36-12.67\%), comparable to those of uniform weighting. The net effect is that precision preservation offsets marginal recall gains, yielding a similar mIoU. Precision and recall performance for DALES are illustrated in Table S4 in the supplementary (Table S6 for RandLA-Net).

S3DIS exhibits more uniform behavior. All methods show moderate rates in both directions (3.7-6.27\% majority-to-minority, 26.82-36.59\% minority-to-majority). Even BalSoft, which achieves the best mIoU (64.85\%), trades lower minority-to-majority (26.82\%) for higher majority-to-minority errors (6.27\%), demonstrating that improved recall is offset by reduced precision. The tighter performance clustering (1.93\% mIoU range) reflects this precision-recall balance across all methods. Precision and recall performance for S3DIS are illustrated in Table S5 in the supplementary (Table S7 for RandLA-Net).

\subsection{Decision Boundary Variability}\label{sec:db_variability}
\begin{figure}[t]
    \includegraphics[width=\textwidth]{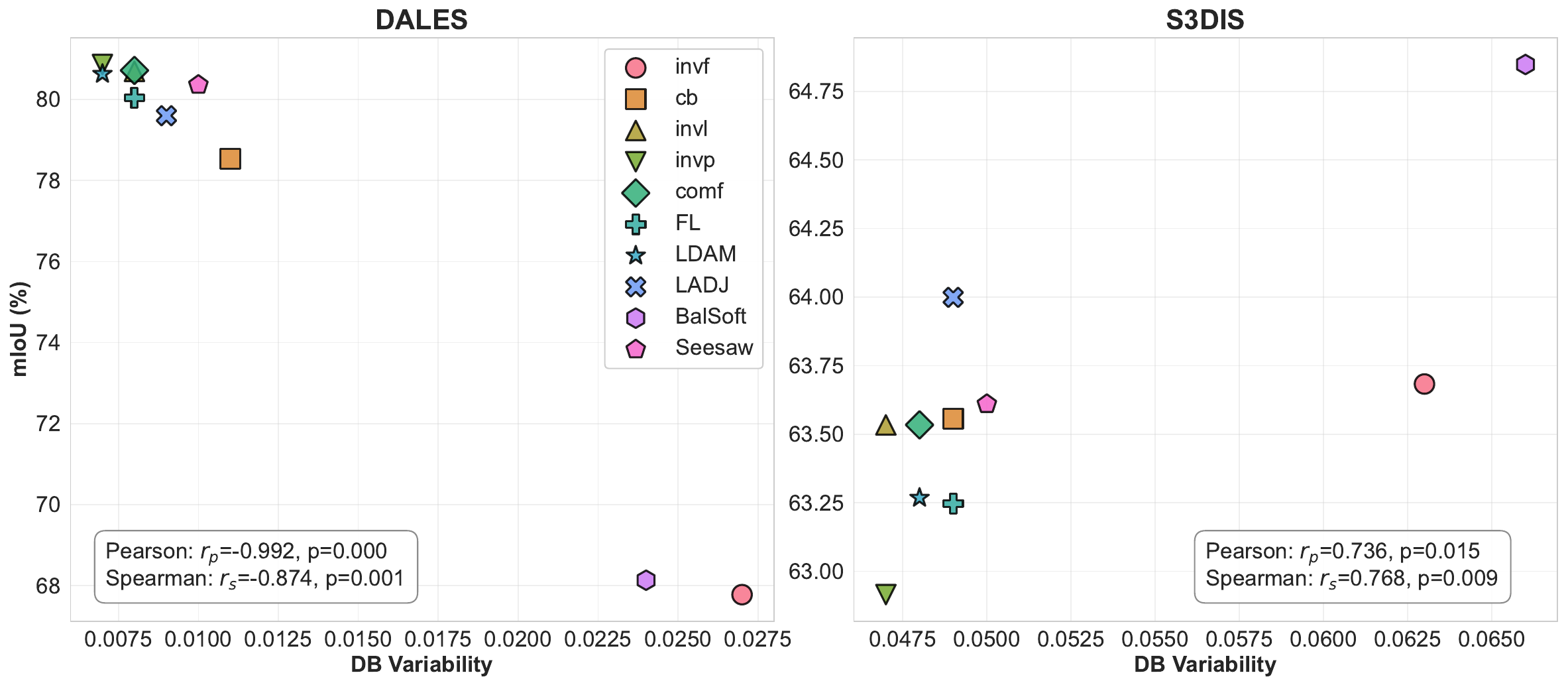}
    \caption{Decision boundary variability vs. mIoU on DALES (left) and S3DIS (right). Each point represents one method; DB variability measures decision boundary divergence from uniform weighting (uni). DALES exhibits a strong negative correlation, with large deviations causing significant performance degradation (invf, BalSoft: >12\% mIoU drop). S3DIS exhibits a weak positive correlation with all methods clustered within $\approx$2\% mIoU.} 
    \label{fig:db_variability}
    \vspace{-0.5cm}
\end{figure}

To understand whether the choice of imbalance mitigation strategy fundamentally alters decision boundaries, we analyze DB variability relative to uniform weighting (Eq.~\eqref{eq:db_variability}). Figure~\ref{fig:db_variability} presents the relationship between DB variability (relative to uni) and mIoU performance for both datasets. On DALES (left), we observe a striking negative correlation with methods exhibiting low DB variability ($<$0.013) achieve 80-81\% mIoU, comparable to or exceeding uni (80.05\%), while methods exhibiting high DB variability ($>$0.020) suffer performance degradation (invf: 67.78\%, BalSoft: 68.14\%). The Spearman correlation is $r_s$=-0.874 (p$<$0.001), indicating that deviating from uni's decision boundaries strongly predicts performance reduction. This pattern suggests that at extreme imbalance (641:1), uni occupies a narrow basin of favorable solutions, with minor deviations (invp, LDAM, invl, comf showcasing DB variability $\approx$ 0.007-0.0088) can provide marginal gains (+0.6-0.8\% mIoU). Still, large deviations risk escaping this basin entirely.

\begin{table}[t!]
\begin{center}
\caption{Analysis of confusion matrices for DALES (left) and S3DIS (right). Values denote the average percentage of points in minority/majority classified in the majority/minority.}\label{tab:confusion_analysis}
    \resizebox{\textwidth}{!}{%
        \begin{tabular}{|c||c|c||c|c|}
            \hline
            method & majority $\rightarrow$ minority & minority $\rightarrow$ majority & majority $\rightarrow$ minority & minority $\rightarrow$ majority \\ \hline
                        
            uni  	& 0.15 & 12.78 & 3.79 & 36.2 \\ \hline
            
            invf	& 1.58 & 1.58  & 5.74 & 32.21 \\
            cb	    & 0.43 & 5.67  & 4.39 & 33.81 \\
            invl	& 0.17 & 11.11 & 4.07 & 35.08 \\
            invp	& 0.19 & 10.36 & 4.04 & 35.66 \\
            comf	& 0.19 & 11.53 & 4.33 & 35.11 \\ \hline
            
            FL	    & 0.17 & 11.82 & 3.7  & 36.59 \\
            LDAM	& 0.15 & 12.67 & 3.97 & 35.8 \\
            LADJ	& 0.34 & 7.35  & 4.57 & 33.03 \\
            BalSoft	& 1.50 & 1.89  & 6.27 & 26.82 \\
            Seesaw	& 0.18 & 11.55 & 4.26 & 35.92 \\ \hline
            
        \end{tabular} %
    }
\end{center}
\vspace{-0.5cm}
\end{table}

In contrast, S3DIS (Fig.~\ref{fig:db_variability} right) exhibits a weak positive correlation ($\rho$=0.768, p=0.009), with all methods clustering within 62.9-64.9\% mIoU despite DB variability ranging from 0.047 to 0.066. The best-performing method (BalSoft: 64.85\%) has the highest DB variability (0.066), while the methods closest to uni (invp, invl) with DB variability $\approx$0.047 do not consistently outperform. 

These contrasting patterns are consistent with dataset-dependent landscape topologies: a narrow favorable basin under extreme imbalance (DALES) versus a broad plateau under moderate imbalance (S3DIS). 

\subsection{Loss Landscape Topology}\label{sec:ll_topology_results}
\begin{figure}[b]
    \centering
    \begin{subfigure}[b]{0.48\textwidth}
        \centering
        \includegraphics[width=\textwidth]{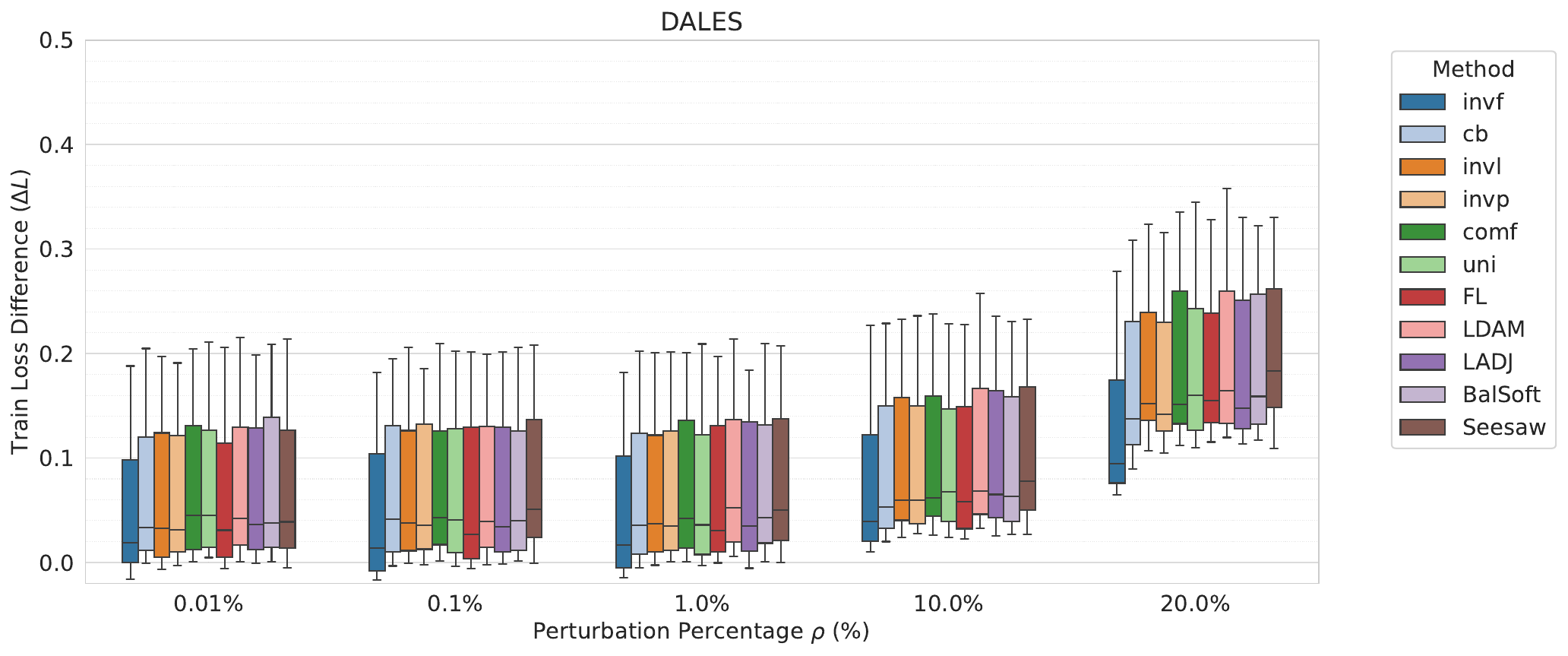}
        \caption{} \label{fig:loss_diff_dales}
    \end{subfigure}
    \hfill
    \begin{subfigure}[b]{0.48\textwidth}
        \centering
        \includegraphics[width=\textwidth]{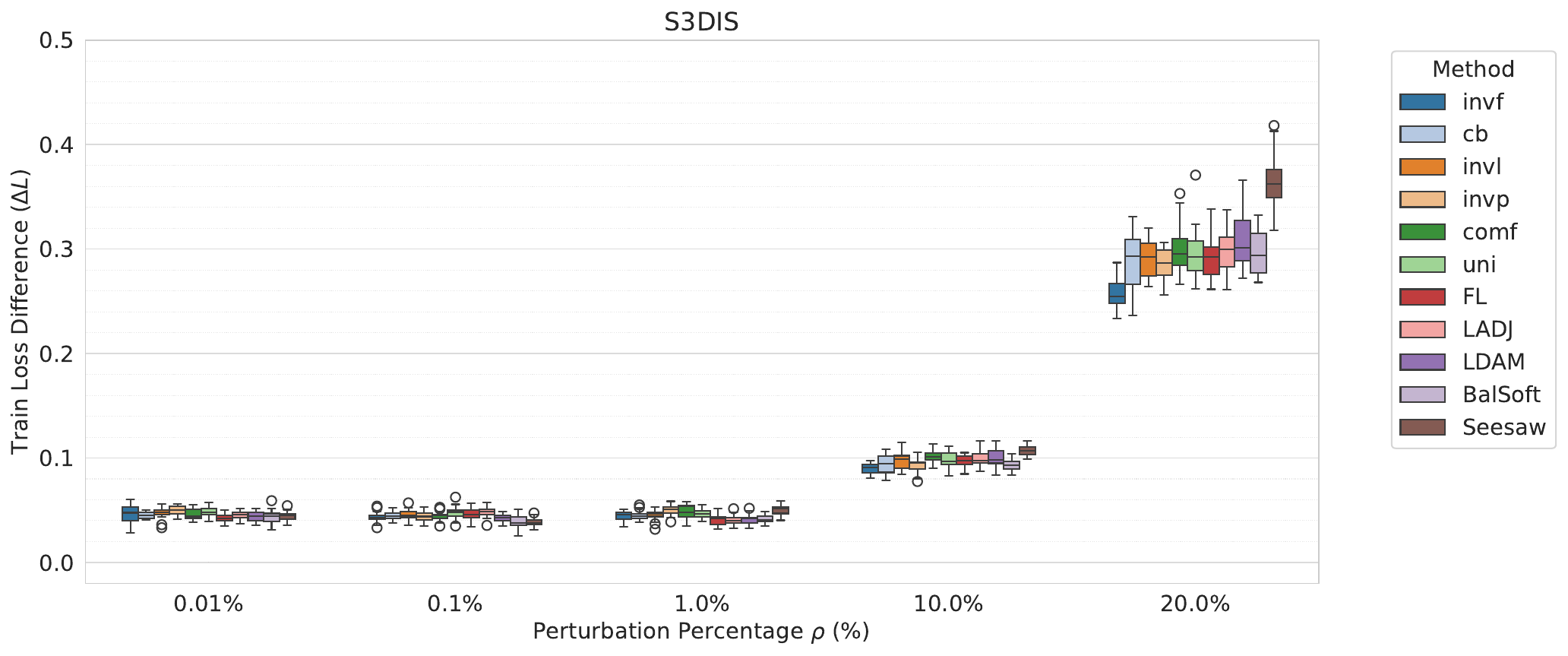}
        \caption{} \label{fig:loss_diff_s3dis}
    \end{subfigure}
    \caption{Train loss differences under weight perturbations (20 random directions). (\subref{fig:loss_diff_dales}) DALES shows moderate sensitivity, while (\subref{fig:loss_diff_s3dis})  S3DIS exhibits flatter landscape. All methods display similar perturbation profiles for each dataset.}
    \label{fig:loss_differences}
    \vspace{-0.5cm}
\end{figure}

To characterize the optimization landscape topology, we perform two complementary analyses: (1) flatness evaluation via perturbation of the network's weights, and (2) Hessian eigenvalue spectrum computation. We evaluate loss sensitivity by sampling K=20 random directions normalized using filter-wise normalization \cite{Li2018}. For perturbation magnitudes $\rho \in \{0.01\%, 0.1\%, 1.0\%, 10\%, 20\%\}$ of the weight norm, we compute train loss differences as in Eq. \eqref{eq:loss_difference}. On DALES (Fig. \ref{fig:loss_diff_dales}), loss differences range from 0 to 0.25 at $\rho$=10\% and 0.08 to 0.35 at $\rho$=20\%, indicating higher sensitivity. On S3DIS (Fig. \ref{fig:loss_diff_s3dis}), $\delta_k$ < 0.02 at $\rho$=10\% and increases to approximately 0.15 at $\rho$=20\%, suggesting a flatter landscape. Notably, all methods exhibit similar perturbation profiles within each dataset, suggesting that the choice of imbalance-aware mitigation strategy does not substantially alter local landscape geometry.

\begin{figure}[t]
    \centering
    \begin{subfigure}[b]{0.48\textwidth}
        \centering
        \includegraphics[width=\textwidth]{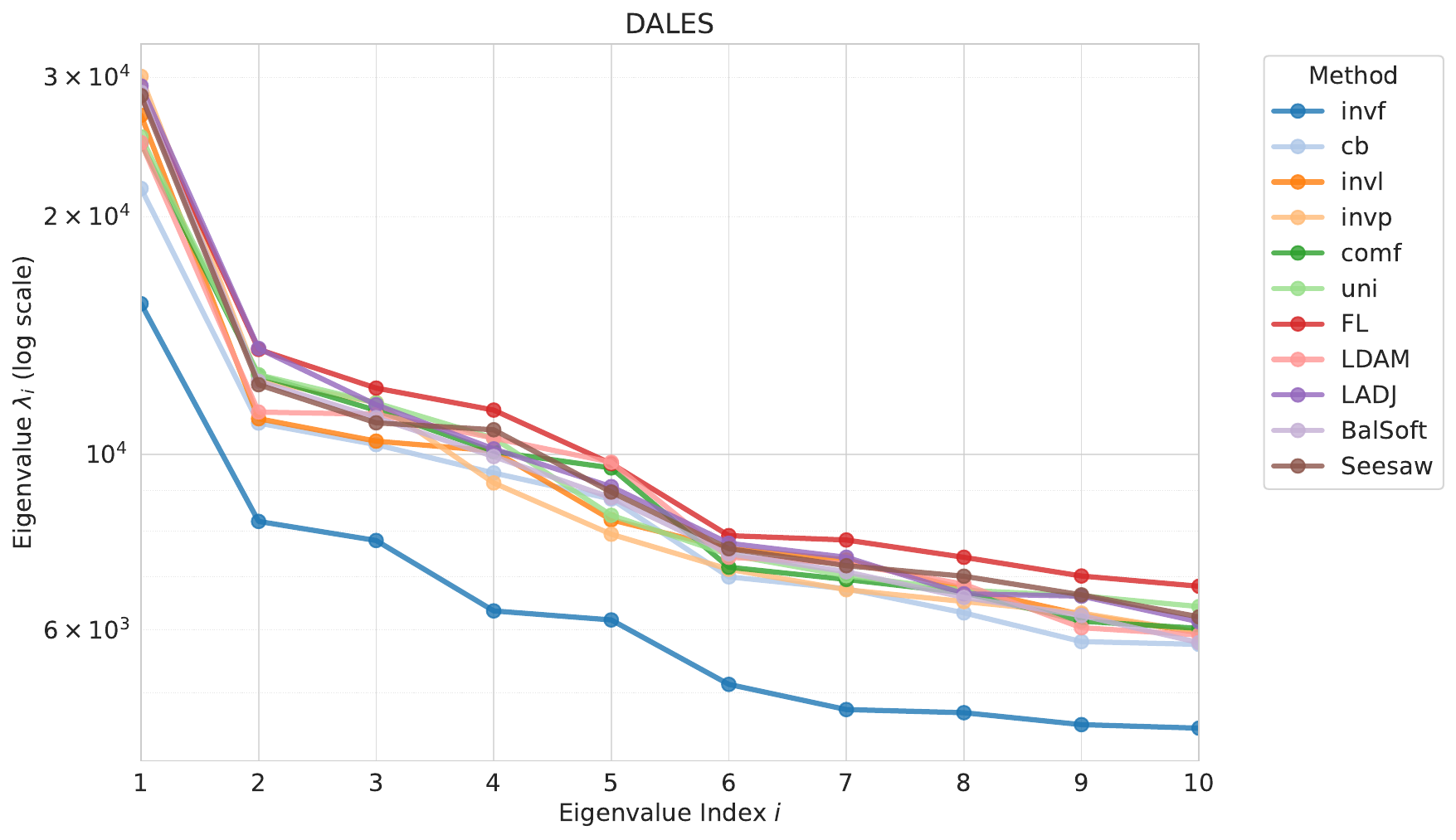}
        \caption{} \label{fig:hessian_dales}
    \end{subfigure}
    \hfill
    \begin{subfigure}[b]{0.48\textwidth}
        \centering
        \includegraphics[width=\textwidth]{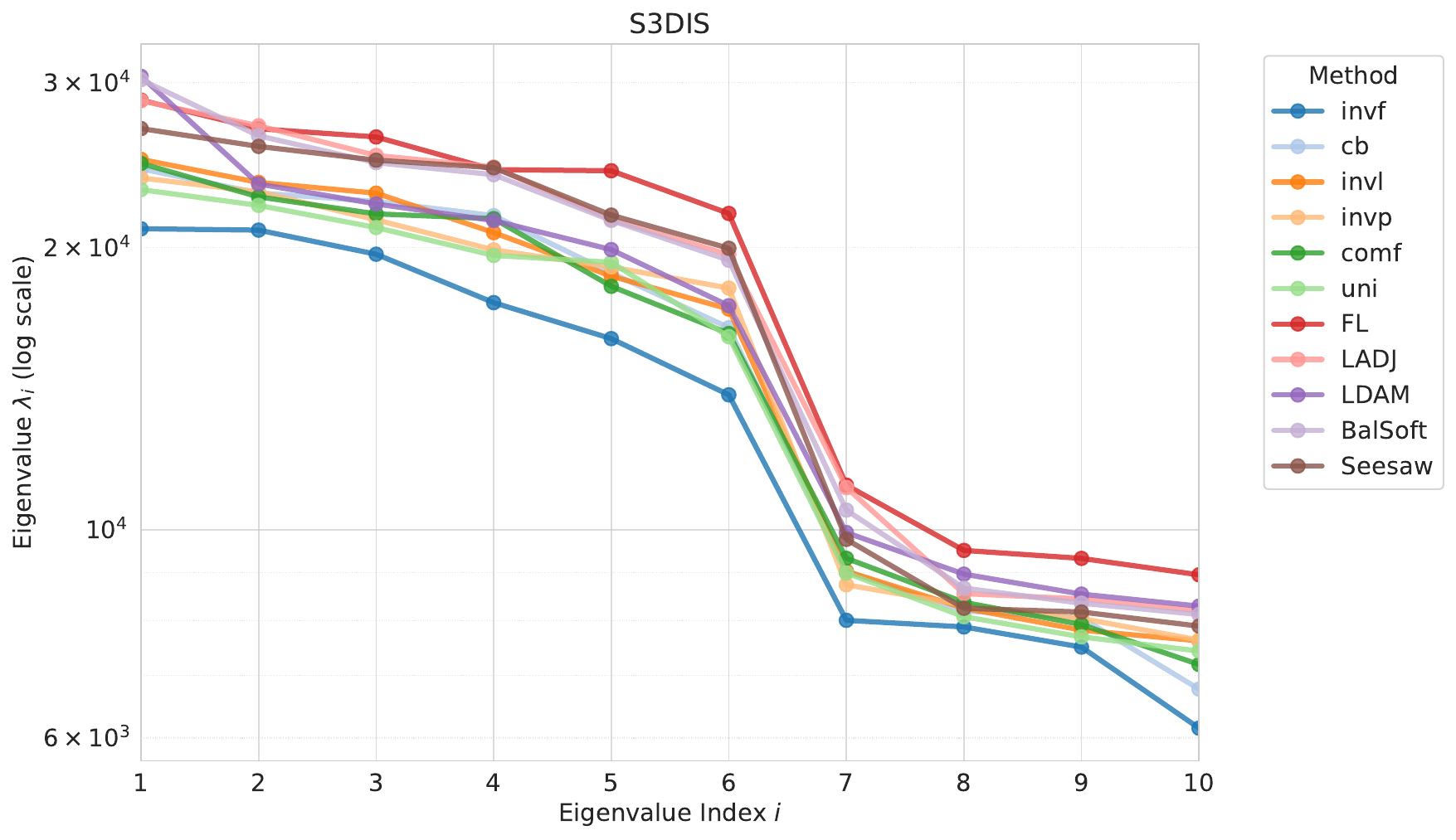}
        \caption{} \label{fig:hessian_s3dis}
    \end{subfigure}
    \caption{Hessian eigenvalue spectra (log scale) for all methods. (\subref{fig:hessian_dales}) DALES exhibits anisotropic curvature (dominant $\lambda_1$), while (\subref{fig:hessian_s3dis}) S3DIS shows more isotropic curvature ($\lambda_1$--$\lambda_6$ similar). Methods cluster tightly, indicating that landscape geometry is determined by data characteristics rather than by the loss formulation.} 
    \label{fig:hessian_spectrum}
    \vspace{-0.5cm}
\end{figure}

We compute the top-10 eigenvalues of the Hessian matrix (Section \ref{sec:ll_topology}; Fig.~\ref{fig:hessian_spectrum}). On DALES, all methods exhibit a dominant first eigenvalue ($\lambda_1$), indicating anisotropic curvature (one sharp direction with relatively flat orthogonal directions). On S3DIS, eigenvalues $\lambda_1$-$\lambda_6$ maintain similar magnitudes before declining, revealing more isotropic curvature. This dataset-level distinction supports our hypothesis of topological differences, i.e., DALES (extreme imbalance) shows a narrow basin, while S3DIS (moderate imbalance) exhibits a flatter plateau.

Critically, eigenvalue spectra are similar across methods within each dataset. This similarity suggests that the choice of loss formulation does not substantially alter the local geometry of converged solutions. Taken together with the decision boundary analysis (Section~\ref{sec:db_variability}), these results indicate that the landscape offers limited degrees of freedom for specialized losses to exploit.

\section{Conclusions and Future Work}\label{sec:conclusions}
Class imbalance mitigation in 3D point cloud segmentation was analyzed, evaluating 11 methods across two point-based architectures on datasets with extreme and moderate imbalance. Our key finding challenges 2D computer vision norms: standard cross-entropy with uniform weighting remains highly competitive, with specialized methods providing modest improvements (0.8-3.3\% mIoU depending on architecture and dataset). Three complementary analyses (precision-recall trade-offs, decision boundary variability, and loss landscape quantification) provide evidence that all methods converge to regions with similar local curvature within each dataset, suggesting that the optimization landscape's topology is primarily shaped by imbalance severity rather than loss formulation.

We hypothesize that this phenomenon stems from differences between 2D and 3D data. Unlike 2D images, which encode semantics through texture and color patterns, 3D point clouds encode information through geometric structure. Point-based architectures exploit this structure through neighborhood aggregation strategies that may provide implicit class balancing, i.e., minority classes are rare in the overall point count but contribute denser local neighborhoods when present. Consequently, explicit loss-level modifications may provide limited additional benefit and can even degrade performance. This architectural-geometric coupling, absent in standard 2D CNNs operating on regular grids, may partly explain why techniques proven effective in 2D do not readily transfer to point-based 3D segmentation.

Validation across KPConv and RandLA-Net with different sampling strategies confirms that the competitive performance of uniform CE persists across architectures, though the magnitude of specialized methods' benefits varies (KPConv: 0.8-1.8\%; RandLA-Net: 2.4-3.3\%). The mechanistic analyses focus on KPConv; extending these analyses to RandLA-Net would strengthen claims about the underlying mechanisms. However, the consistency of performance patterns across both evaluated architectures suggests our core finding, that dataset characteristics constrain the effectiveness of loss-level modifications, is not KPConv-specific. Further validation with voxel-based (e.g., Cylinder3D) and attention-based (Point Transformers) architectures is needed to determine whether these conclusions generalize beyond point-based methods.

\bibliographystyle{splncs04}
\bibliography{refs}

\clearpage

\appendix

\renewcommand{\thesection}{S\arabic{section}}
\setcounter{figure}{0}
\renewcommand{\thefigure}{S\arabic{figure}}
\setcounter{table}{0}
\renewcommand{\thetable}{S\arabic{table}}

\setcounter{page}{1}

\begin{center}
    {\Large \textbf{Loss Landscape Topology Reveals Why Simple Baselines are Competitive at 3D Point Cloud Segmentation Under Class Imbalance} \\
    \vspace{0.5em}Supplementary Material \\
    \vspace{1.0em}}
\end{center}

\section{Network Architectures and Training Protocols}\label{sec-supl:net-train}
\begin{figure}[b]
    \centering
    \begin{subfigure}[b]{0.48\textwidth}
        \centering
        \includegraphics[width=\textwidth]{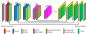}
        \caption{} \label{fig:kpconv_dales}
    \end{subfigure}
    \hfill
    \begin{subfigure}[b]{0.48\textwidth}
        \centering
        \includegraphics[width=\textwidth]{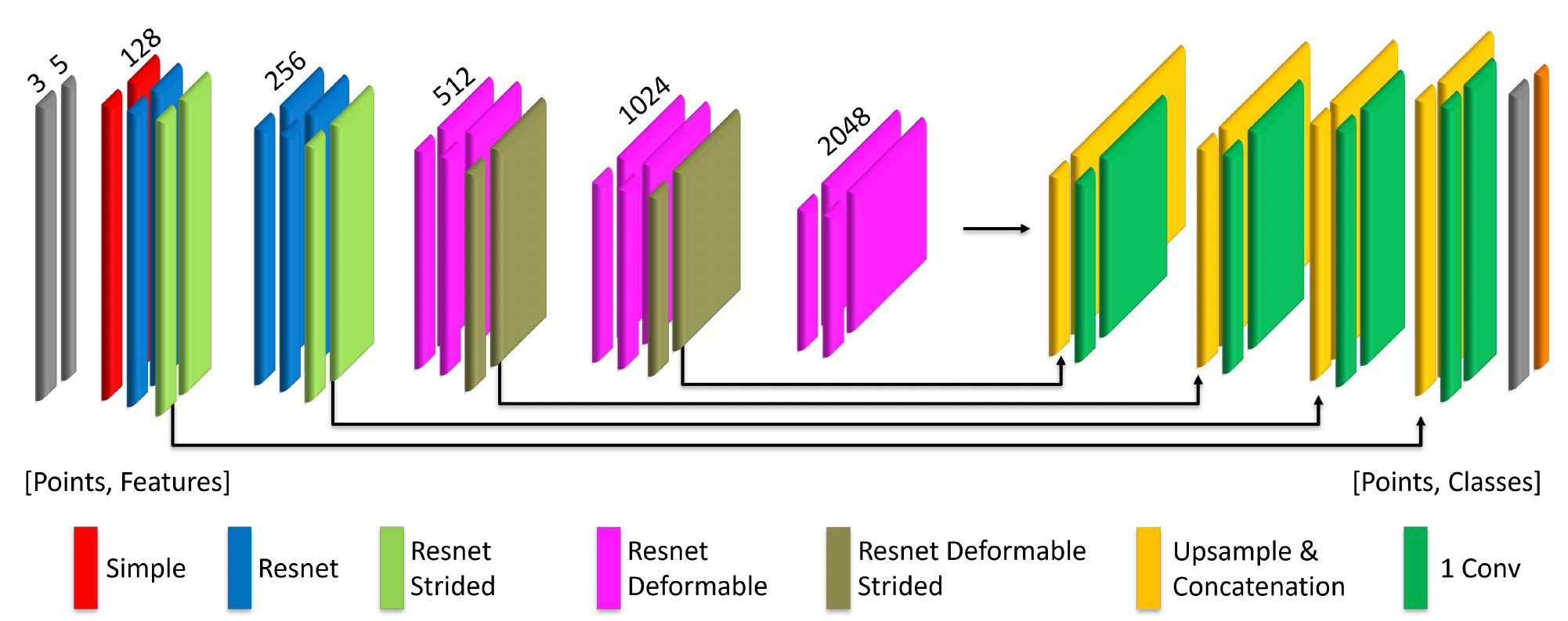}
        \caption{} \label{fig:kpconv_s3dis}
    \end{subfigure}
    \caption{KPConv architectures with dataset-specific configurations for: (\subref{fig:kpconv_dales}) DALES (XYZ features), and (\subref{fig:kpconv_s3dis}) S3DIS (XYZ+RGB features).} \label{fig:kpconv_archs}
\end{figure}

KPConv processes unstructured point clouds directly via a hierarchical encoder-decoder architecture with skip connections. Figure \ref{fig:kpconv_archs} illustrates the specific architectures used, following configurations from \cite{Thomas2019,Varney2020}. For training, we used the network parameters outlined in \cite{Thomas2019}, i.e., a momentum gradient descent optimizer with a momentum value of 0.02, and an initial learning rate of 0.01. The learning rate schedule remained consistent, and dropout was not applied. For DALES, the inputs are $(X,Y,Z)$ coordinates, with a batch size of 10 and 400 epochs, while for S3DIS, we additionally use RGB features, with a batch size of 6 and 500 epochs. In both cases, each epoch consists of 500 optimizer steps. In each case, training was performed using a fixed seed (42) for reproducibility, on a single NVIDIA GPU.

RandLA-Net processes point clouds using stacked local-feature aggregation modules with progressive random sampling, following the architecture from \cite{Hu2020RandLANet}. The network uses five layers with sub-sampling ratios [4, 4, 4, 4, 2] and feature dimensions [16, 64, 128, 256, 512], aggregating K=16 nearest neighbors at each layer. For training, we used the Adam optimizer with an initial learning rate of 0.01, which was reduced by 5\% after each epoch, following \cite{Hu2020RandLANet}. For DALES, the inputs are $(X, Y, Z)$ coordinates, whereas for S3DIS, we also use RGB features. Both datasets use batch size 6, 100 epochs, and 40,960 input points per sample, with each epoch consisting of 500 optimizer steps. Training was performed on a single NVIDIA GPU.

\section{Datasets and Class Distributions}\label{sec-supl:data}

\subsection{DALES}
Consists of outdoor aerial LiDAR covering 40 urban/rural tiles (0.5 km$^2$ each) with eight semantic classes (excluding \textit{unknown}): \textit{ground}, \textit{vegetation}, \textit{cars}, \textit{trucks}, \textit{power lines}, \textit{fences}, \textit{poles}, and \textit{buildings}. The dataset exhibits extreme imbalance with a 641:1 ratio between the most frequent (\textit{ground}: 48.4\%) and least frequent (\textit{poles}: 0.08\%) classes \cite{Varney2020} (Fig. \ref{fig:distr_dales} of supplementary). One tile was excluded from the training set due to labelling issues (the majority of points were annotated as \textit{unknown}).

\subsection{S3DIS}
Consists of indoor scans from 6 large-scale building areas with 13 semantic classes, including structural elements (\textit{ceiling}, \textit{floor}, \textit{wall}), architectural components (\textit{beam}, \textit{column}, \textit{window}, \textit{door}), and furniture (\textit{chair}, \textit{table}, \textit{bookcase}, \textit{sofa}, \textit{board}, \textit{clutter}). This dataset has a moderate imbalance ratio (56:1; most frequent is \textit{ceiling} (19.14\%) and least frequent is \textit{sofa}: (0.49\%)) \cite{Armeni2016} (Fig. \ref{fig:distr_s3dis} of supplementary).

\subsection{Class Distributions}
Figure~\ref{fig:data_distribution} shows the logarithmic class distributions for DALES (643:1 imbalance) and S3DIS (55:1 imbalance), exhibiting characteristic long-tailed patterns where majority classes dominate point counts.

\begin{figure}[t]
    \centering
    \begin{subfigure}[b]{0.48\textwidth}
        \centering
        \includegraphics[width=\textwidth]{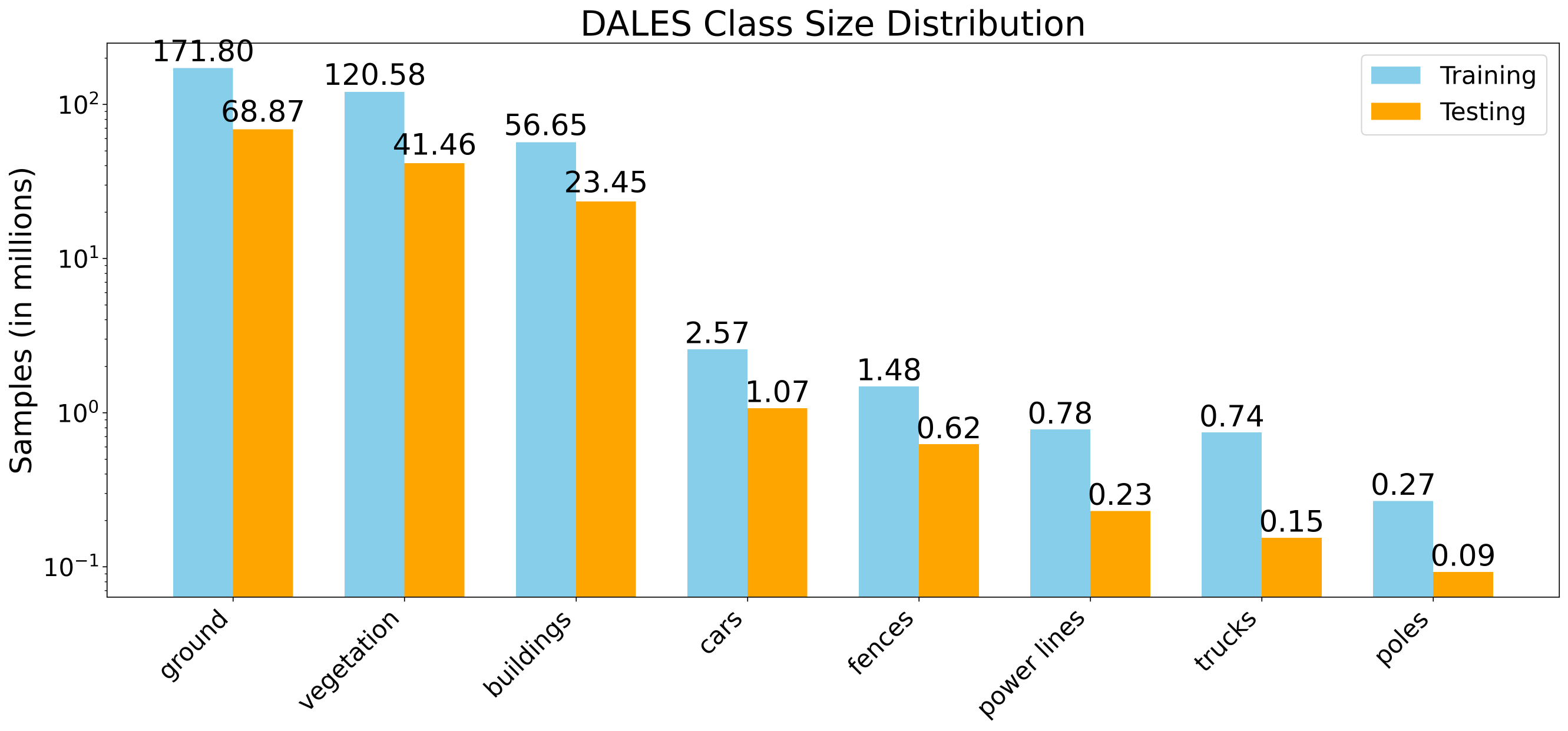}
        \caption{} \label{fig:distr_dales}
    \end{subfigure}
    \hfill
    \begin{subfigure}[b]{0.48\textwidth}
        \centering
        \includegraphics[width=\textwidth]{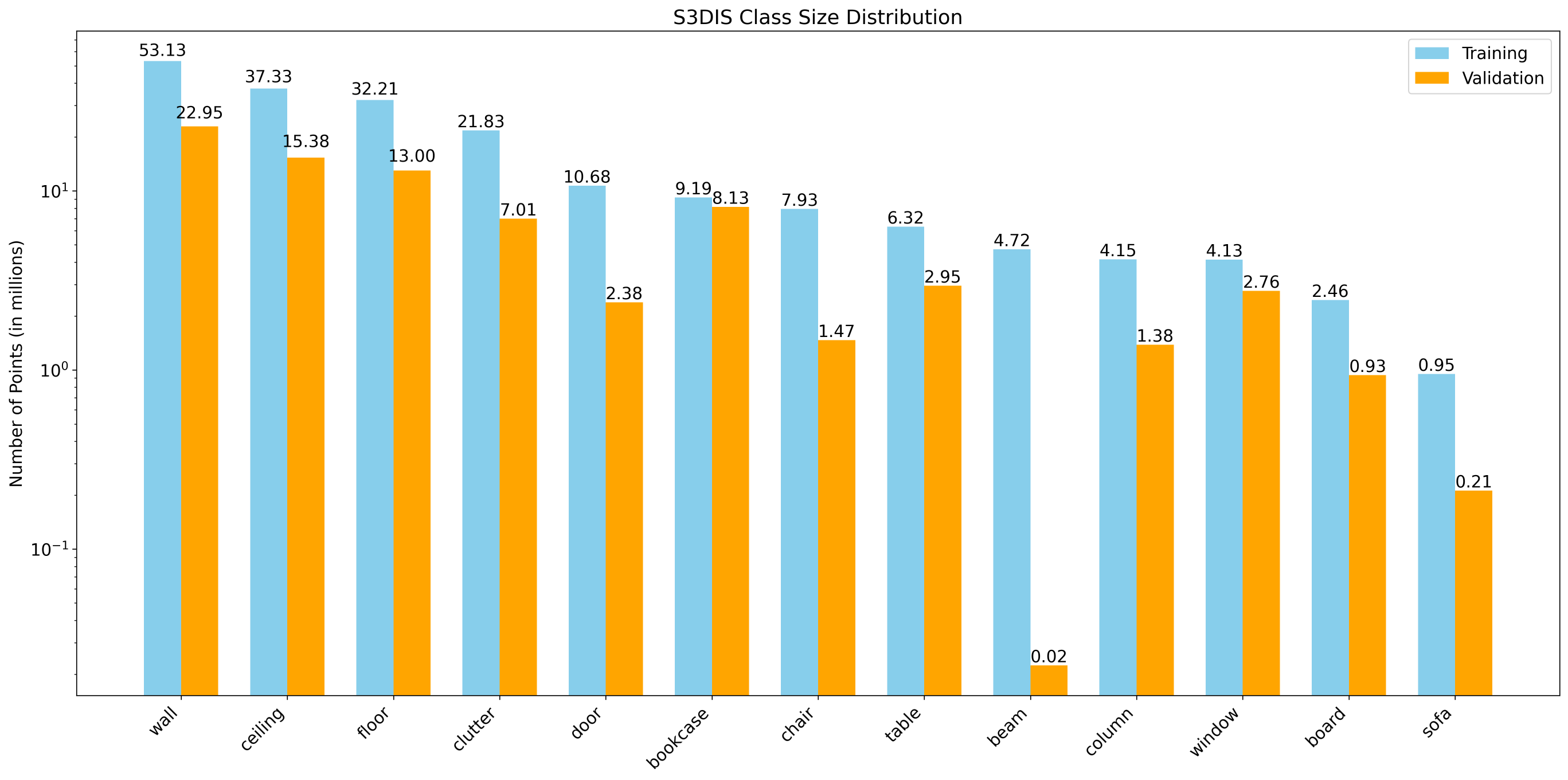}
        \caption{} \label{fig:distr_s3dis}
    \end{subfigure}
    \caption{Logarithmic class distributions showing long-tailed characteristics with: (\subref{fig:distr_dales}) extreme (DALES) and (\subref{fig:distr_s3dis}) moderate (S3DIS) imbalance ratios.} \label{fig:data_distribution}
\end{figure}

\section{Seed-Based Variability Analysis (with KPConv)}\label{sec:seed_based_var}
To ensure the observed patterns are not artifacts of random seed selection, we trained three additional runs (using different seeds from Section~\ref{sec:overall_perf}, i.e., seeds 1, 2, and 3) for uniform weighting (uni) and the best-performing method on each dataset (invp for DALES, BalSoft for S3DIS). Table~\ref{tab:seeds_iou} presents $\mu\pm\sigma$ across these runs. On DALES, uniform weighting achieves $80.01\pm0.17\%$, while invp achieves $81\pm0.1\%$, confirming the competitive performance and low variance of both methods. Notably, the performance gap (0.99\%) is larger than the original single-seed result (0.81\%), and standard deviations are minor (0.10-0.17\%), indicating stable convergence. Per-class standard deviation is highest for minority classes (\textit{trucks}: 0.61\%; fences: 0.80\% for uni), yet both methods maintain consistent rankings across seeds.

On S3DIS, uniform weighting achieves $63.54\pm0.2\%$, while BalSoft achieves $64.92\pm0.27\%$, confirming the tight clustering observed in single-seed results (1.38\% gap vs. 1.75\% initially). Standard deviations remain small across both methods (0.2-0.37\%), with minority classes again showing higher variance (\textit{column}: 2.55\%; \textit{sofa}: 2.71\% for uni). Critically, the ranking stability persists, with BalSoft being marginally better than uni across all seeds.

\begin{table}[t]
\begin{flushleft}
\caption{Seed-based variability analysis across three runs (different seeds, i.e., 1, 2, and 3) $(\mu~\pm~\sigma)$.}\label{tab:seeds_iou}
    
    \vspace{5pt}
    \textbf{(a) DALES Dataset}
    \vspace{2pt}
    
    \scriptsize
    \begin{tabular}{|l||c|c|c|c|c|}
    \hline
    Method & mean & \textit{ground} & \textit{vegetation} & \textit{cars} & \textit{trucks}\\
    \hline
    \hline
    uni     & 80.005 $\pm$ 0.167 & 96.577 $\pm$ 0.027 & 93.796 $\pm$ 0.036 & 84.978 $\pm$ 0.159 & 43.046 $\pm$ 0.608 \\
    invp    & 80.998 $\pm$ 0.097 & 96.477 $\pm$ 0.094 & 93.808 $\pm$ 0.037 & 85.152 $\pm$ 0.197 & 45.725 $\pm$ 0.515 \\
    \hline
    \end{tabular}  

    \vspace{1pt}

    \scriptsize
    \begin{tabular}{|l||c|c|c|c|}
    \hline
    Method & \textit{power lines} & \textit{fences} & \textit{poles} & \textit{buildings} \\
    \hline 
    \hline
    uni     & 93.775 $\pm$ 0.318 & 60.907 $\pm$ 0.798 & 71.830 $\pm$ 0.355 & 95.130 $\pm$ 0.098 \\
    invp    & 94.530 $\pm$ 0.222 & 63.141 $\pm$ 0.263 & 74.305 $\pm$ 0.449 & 94.848 $\pm$ 0.306 \\
    \hline
    \end{tabular}
    
    \vspace{5pt}
    \textbf{(b) S3DIS Dataset}
    \vspace{2pt}

    \resizebox{\textwidth}{!}{%
        \begin{tabular}{|l||c|c|c|c|c|c|c|}
        \hline
         Method & mean & \textit{ceiling} & \textit{floor} & \textit{wall} & \textit{beam} & \textit{column} & \textit{window} \\
        \hline
        \hline
        uni     & 63.541 $\pm$ 0.196 & 93.196 $\pm$ 0.152 & 98.479 $\pm$ 0.021 & 80.604 $\pm$ 0.180 & 0.000 $\pm$ 0.000 & 23.002 $\pm$ 2.546 & 45.451 $\pm$ 1.209 \\
        BalSoft & 64.919 $\pm$ 0.270 & 92.586 $\pm$ 0.359 & 98.389 $\pm$ 0.087 & 82.038 $\pm$ 0.388 & 0.003 $\pm$ 0.005 & 29.711 $\pm$ 0.852 & 53.546 $\pm$ 1.318 \\
        \hline
        \end{tabular}%
    }

    \vspace{1pt}
    
    \resizebox{\textwidth}{!}{%
        \begin{tabular}{|l||c|c|c|c|c|c|c|}
        \hline
         Method & \textit{door} & \textit{chair} & \textit{table} & \textit{bookcase} & \textit{sofa} & \textit{board} & \textit{clutter} \\
        \hline
        uni     & 62.568 $\pm$ 0.098 & 87.831 $\pm$ 0.717 & 79.076 $\pm$ 0.520 & 70.903 $\pm$ 0.514 & 65.376 $\pm$ 2.709 & 61.845 $\pm$ 1.139 & 57.696 $\pm$ 1.019 \\
        BalSoft & 63.998 $\pm$ 1.329 & 87.062 $\pm$ 0.070 & 78.108 $\pm$ 0.653 & 71.214 $\pm$ 0.332 & 67.067 $\pm$ 1.447 & 63.323 $\pm$ 0.779 & 56.904 $\pm$ 0.732 \\
        \hline
        \end{tabular}%
    }
\end{flushleft}
\vspace{-0.6cm}
\end{table}

\section{Overall Performance (with RandLA-Net)}

We also validated findings using RandLA-Net \cite{Hu2020RandLANet}, which employs fundamentally different design principles, i.e., random point sampling vs. structured potential-based sampling as in KPConv. This comparison tests whether findings stem from KPConv-specific properties or reflect general characteristics of point-based 3D segmentation. Tables \ref{tab:full_iou_dales_randla}-\ref{tab:full_iou_S3DIS_randla} present complete per-class IoU results for DALES and S3DIS, respectively.

\begin{table}[t]
\begin{center}
\vspace{-0.5cm}
\caption{Per-class performance on the DALES dataset using RandLA-Net.}\label{tab:full_iou_dales_randla}
    \resizebox{\textwidth}{!}{%
        \begin{tabular}{|l||c|c|c|c|c|c|c|c|c|}

        \hline
        Method & mean & \textit{ground} & \textit{vegetation} & \textit{cars} & \textit{trucks} & \textit{power lines} & \textit{fences} & \textit{poles} & \textit{buildings} \\ 

        \hline
        \hline
        uni     & 76.762 & 97.148 & 93.463 & 83.352 & 38.220 & 91.493 & 53.511 & 60.266 & 96.641 \\
        
        \hline
        invf    & 67.372 & 96.313 & 89.435 & 64.182 & 32.725 & 91.212 & 23.269 & 46.648 & 95.193 \\
        cb      & 77.068 & 97.007 & 92.955 & 80.660 & 37.879 & 93.530 & 53.537 & 64.579 & 96.401 \\
        invl    & 76.843 & 97.034 & 93.175 & 83.681 & 39.589 & 89.842 & 54.049 & 60.650 & 96.726 \\
        invp    & 78.481 & 97.109 & 93.499 & 83.569 & 39.585 & 93.032 & 56.368 & 68.105 & 96.577 \\
        comf    & 76.960 & 97.035 & 93.232 & 83.347 & 39.306 & 91.040 & 54.002 & 60.994 & 96.728 \\

        \hline
        FL	    & 75.972 & 97.155 & 93.356 & 83.481 & 32.884 & 90.917 & 52.261 & 61.089 & 96.633 \\
        LDAM	& 79.172 & 97.182 & 93.603 & 84.144 & 38.690 & 93.199 & 57.326 & 72.420 & 96.809 \\
        LADJ	& 66.762 & 97.038 & 91.130 & 73.658 & 18.631 & 90.722 & 25.906 & 40.825 & 96.186 \\
        BalSoft	& 66.430 & 96.852 & 90.782 & 74.525 & 24.390 & 88.973 & 24.881 & 34.702 & 96.333 \\
        Seesaw	& 74.260 & 97.192 & 92.805 & 79.831 & 30.681 & 91.024 & 45.357 & 60.493 & 96.694 \\
       
        \hline \hline
        Range & 12.742 & 0.879 & 4.168 & 19.962 & 20.958 & 4.557 & 34.057 & 37.718 & 1.616 \\

        \hline
        \end{tabular}%
    }
\end{center}
\end{table}

\begin{table}[ht]
\begin{center}
\vspace{-0.5cm}
\caption{Per-class performance on the S3DIS dataset using RandLA-Net.}\label{tab:full_iou_S3DIS_randla}
    \resizebox{\textwidth}{!}{%
        \begin{tabular}{|l||c|c|c|c|c|c|c|c|c|c|c|c|c|c|}

        \hline
        Method & mean & \textit{ceiling} & \textit{floor} & \textit{wall} & \textit{beam} & \textit{column} & \textit{window} & \textit{door} & \textit{chair} & \textit{table} & \textit{bookcase} & \textit{sofa} & \textit{board} & \textit{clutter} \\

        \hline
        \hline
        uni     & 61.375 & 93.116 & 97.028 & 80.367 & 0.000 & 17.390 & 57.660 & 36.838 & 78.216 & 84.665 & 55.844 & 70.799 & 71.421 & 54.524 \\

        \hline
        invf	& 60.373 & 91.363 & 97.390 & 78.549 & 0.000 & 15.893 & 60.357 & 30.560 & 76.231 & 85.703 & 61.124 & 71.363 & 64.896 & 51.420 \\
        cb	    & 62.406 & 91.821 & 97.074 & 79.904 & 0.000 & 26.317 & 61.150 & 33.409 & 78.371 & 81.896 & 75.646 & 70.521 & 64.509 & 50.663 \\
        invl	& 61.661 & 92.351 & 97.355 & 80.518 & 0.000 & 16.287 & 59.860 & 39.551 & 77.654 & 86.433 & 60.295 & 70.575 & 68.157 & 52.555 \\
        invp	& 63.686 & 92.722 & 97.734 & 80.037 & 0.000 & 22.035 & 59.819 & 41.352 & 78.766 & 86.838 & 72.210 & 71.187 & 72.921 & 52.304 \\
        comf	& 62.640 & 92.319 & 97.902 & 80.974 & 0.000 & 24.176 & 59.874 & 33.803 & 78.060 & 87.450 & 61.501 & 71.115 & 73.440 & 53.712 \\

        \hline
        FL	    & 62.526 & 91.577 & 97.346 & 81.361 & 0.000 & 21.368 & 58.047 & 52.706 & 76.267 & 86.805 & 55.145 & 71.403 & 68.610 & 52.198 \\
        LDAM	& 64.698 & 92.158 & 96.902 & 81.553 & 0.000 & 28.788 & 59.396 & 50.455 & 77.356 & 88.658 & 66.383 & 72.024 & 73.208 & 54.194 \\
        LADJ	& 64.669 & 92.080 & 96.661 & 82.223 & 0.000 & 32.908 & 62.755 & 47.018 & 75.976 & 88.004 & 69.933 & 72.854 & 67.595 & 52.687 \\
        BalSoft	& 63.117 & 91.241 & 97.327 & 81.789 & 0.000 & 23.710 & 61.296 & 43.184 & 78.355 & 87.483 & 67.940 & 71.363 & 63.855 & 52.973 \\
        Seesaw	& 62.491 & 91.811 & 97.823 & 80.882 & 0.000 & 20.586 & 59.820 & 45.444 & 78.221 & 86.209 & 56.853 & 71.604 & 69.960 & 53.166 \\

        \hline \hline
        Range & 4.325 & 1.875 & 1.241 & 3.674 & 0.000 & 17.015 & 5.095 & 22.146 & 2.790 & 6.762 & 20.501 & 2.333 & 9.585 & 3.861 \\

        \hline
        \end{tabular}%
    }
\end{center}
\end{table}

\newpage
\section{Precision-Recall Patterns}

\begin{table}[t]
\begin{center}
\vspace{-0.8cm}
\caption{Precision (upper part) and recall (lower part) on DALES dataset using KPConv.} \label{tab:precision_recall_dales}
    \resizebox{\textwidth}{!}{%
        \begin{tabular}{|l||c|c|c|c|c|c|c|c|c|}

        \hline
        Method & mean & \textit{ground} & \textit{vegetation} & \textit{cars} & \textit{trucks} & \textit{power lines} & \textit{fences} & \textit{poles} & \textit{buildings} \\
        \hline
        \hline
        \multicolumn{10}{c}{Precision} \\
        \hline
        uni   	& 89.810 & 97.494 & 97.389 & 90.898 & 66.322 & 97.442 & 84.588 & 86.107 & 98.244 \\
        \hline
        invf	& 70.696 & 97.793 & 97.765 & 74.209 & 37.779 & 88.948 & 31.534 & 40.854 & 96.688 \\
        cb	    & 83.462 & 97.447 & 97.691 & 90.496 & 58.262 & 97.351 & 62.021 & 66.010 & 98.421 \\
        invl	& 90.378 & 97.432 & 97.671 & 89.148 & 74.008 & 97.538 & 82.557 & 86.632 & 98.042 \\
        invp	& 89.336 & 97.456 & 97.607 & 89.937 & 68.918 & 97.659 & 79.673 & 85.183 & 98.255 \\
        comf	& 89.965 & 97.474 & 97.429 & 90.761 & 72.899 & 97.632 & 77.497 & 87.633 & 98.392 \\
		\hline							
        FL	    & 89.757 & 97.236 & 97.703 & 89.799 & 70.484 & 96.671 & 81.825 & 85.759 & 98.577 \\
        LDAM	& 90.822 & 97.405 & 97.498 & 90.906 & 74.017 & 97.321 & 82.987 & 88.010 & 98.435 \\
        LADJ	& 85.702 & 97.501 & 97.702 & 88.581 & 57.605 & 97.596 & 68.139 & 80.036 & 98.458 \\
        BalSoft	& 70.964 & 97.885 & 98.063 & 78.068 & 25.069 & 97.667 & 34.497 & 39.839 & 96.624 \\
        Seesaw	& 89.690 & 97.455 & 97.705 & 90.208 & 71.743 & 97.283 & 79.553 & 85.482 & 98.090 \\

        \hline \hline
        \multicolumn{10}{c}{Recall} \\
        \hline
        uni	    & 85.773 & 98.961 & 96.199 & 93.078 & 54.441 & 96.409 & 68.744 & 81.800 & 96.553 \\
        \hline
        invf	& 92.593 & 98.006 & 93.049 & 95.203 & 69.793 & 96.798 & 95.206 & 95.954 & 96.735 \\
        cb	    & 90.742 & 98.945 & 95.586 & 93.456 & 63.150 & 97.271 & 87.731 & 93.514 & 96.281 \\
        invl	& 86.415 & 99.006 & 95.995 & 94.597 & 51.350 & 96.784 & 72.220 & 84.734 & 96.631 \\
        invp	& 87.330 & 99.012 & 96.003 & 94.101 & 55.475 & 96.844 & 75.306 & 85.326 & 96.573 \\
        comf	& 86.695 & 98.947 & 96.137 & 93.281 & 51.327 & 96.870 & 77.525 & 83.046 & 96.427 \\
        \hline
        FL	    & 85.964 & 99.216 & 95.881 & 94.115 & 52.948 & 96.921 & 72.443 & 79.988 & 96.202 \\
        LDAM	& 85.914 & 99.057 & 96.142 & 93.210 & 52.809 & 97.034 & 71.716 & 80.931 & 96.413 \\
        LADJ	& 89.118 & 99.011 & 95.775 & 94.312 & 61.546 & 96.518 & 83.126 & 86.174 & 96.483 \\
        BalSoft	& 92.006 & 98.165 & 93.465 & 94.352 & 70.000 & 95.900 & 93.257 & 94.030 & 96.880 \\
        Seesaw	& 86.462 & 99.035 & 95.941 & 93.750 & 52.699 & 96.524 & 75.149 & 81.882 & 96.712 \\

        \hline \hline
        \end{tabular}%
    }
\end{center}
\vspace{-0.5cm}
\end{table}

\begin{table}[h]
\begin{center}
\vspace{-0.5cm}
\caption{Precision (upper part) and recall (lower part) on S3DIS dataset using KPConv.} \label{tab:precision_recall_s3dis}
    \resizebox{\textwidth}{!}{%
        \begin{tabular}{|l||c|c|c|c|c|c|c|c|c|c|c|c|c|c|}
        \hline
        Method & mean & \textit{ceiling} & \textit{floor} & \textit{wall} & \textit{beam} & \textit{column} & \textit{window} & \textit{door} & \textit{chair} & \textit{table} & \textit{bookcase} & \textit{sofa} & \textit{board} & \textit{clutter} \\
        \hline
        \hline
        \multicolumn{15}{c}{Precision} \\
        \hline
        uni	    & 81.565 & 97.293 & 99.189 & 83.317 & 0.000 & 80.802 & 90.412 & 79.342 & 92.464 & 88.386 & 87.970 & 95.393 & 93.337 & 72.436 \\
        \hline
        invf	& 78.439 & 97.256 & 99.218 & 85.021 & 0.000 & 63.928 & 87.944 & 75.279 & 91.147 & 85.560 & 84.813 & 92.295 & 85.934 & 71.316 \\
        cb	    & 79.976 & 97.204 & 99.156 & 83.692 & 0.000 & 70.435 & 88.512 & 79.108 & 91.532 & 85.561 & 87.193 & 93.828 & 91.461 & 72.001 \\
        invl	& 81.193 & 97.514 & 99.208 & 83.270 & 0.000 & 77.553 & 92.832 & 78.018 & 92.147 & 88.147 & 87.545 & 93.591 & 91.378 & 74.304 \\
        invp	& 81.408 & 97.384 & 99.080 & 82.839 & 0.000 & 80.123 & 91.607 & 79.555 & 91.734 & 87.357 & 87.349 & 94.456 & 93.360 & 73.464 \\
        comf	& 81.397 & 97.104 & 99.111 & 83.800 & 0.000 & 79.199 & 90.457 & 82.339 & 91.315 & 88.350 & 86.817 & 94.667 & 92.802 & 72.201 \\
        \hline
        FL	    & 81.557 & 97.288 & 99.040 & 82.106 & 0.000 & 77.596 & 91.282 & 80.918 & 92.836 & 88.108 & 88.129 & 96.084 & 95.333 & 71.526 \\
        LDAM	& 81.206 & 97.167 & 99.181 & 83.224 & 0.000 & 75.858 & 91.819 & 81.969 & 92.589 & 88.253 & 87.420 & 94.464 & 92.174 & 71.559 \\
        LADJ	& 80.581 & 97.421 & 99.173 & 84.304 & 0.000 & 75.538 & 88.741 & 79.821 & 91.358 & 87.319 & 86.883 & 92.672 & 91.490 & 72.828 \\
        BalSoft	& 76.355 & 97.537 & 99.348 & 88.562 & 0.000 & 61.114 & 82.040 & 76.116 & 90.521 & 83.663 & 84.620 & 79.317 & 78.356 & 71.422 \\
        Seesaw	& 81.308 & 97.256 & 99.147 & 83.407 & 0.000 & 75.522 & 91.293 & 83.615 & 92.561 & 87.675 & 87.163 & 95.109 & 93.637 & 70.614 \\
        
        \hline \hline
        \multicolumn{15}{c}{Recall} \\
        \hline
        uni	    & 69.134 & 96.201 & 99.314 & 96.483 & 0.000 & 22.481 & 46.081 & 73.363 & 94.041 & 88.227 & 78.755 & 67.759 & 62.205 & 73.838 \\
        \hline
        invf	& 71.095 & 94.730 & 99.082 & 93.528 & 0.000 & 29.803 & 49.728 & 75.432 & 95.432 & 91.865 & 81.376 & 75.961 & 66.882 & 70.411 \\
        cb	    & 70.232 & 95.186 & 99.222 & 95.475 & 0.000 & 27.048 & 47.916 & 74.670 & 95.090 & 90.147 & 79.131 & 71.002 & 66.177 & 71.952 \\
        invl	& 69.724 & 95.730 & 99.279 & 96.382 & 0.000 & 25.449 & 47.205 & 71.048 & 94.747 & 88.921 & 79.968 & 70.248 & 63.879 & 73.549 \\
        invp	& 69.039 & 95.377 & 99.323 & 96.608 & 0.000 & 24.926 & 48.525 & 71.479 & 94.240 & 89.012 & 78.247 & 63.073 & 64.182 & 72.517 \\
        comf	& 69.709 & 95.586 & 99.274 & 96.135 & 0.000 & 22.856 & 48.844 & 77.649 & 95.302 & 88.819 & 79.473 & 65.249 & 64.867 & 72.158 \\
        \hline
        FL	    & 69.192 & 94.647 & 99.341 & 96.455 & 0.000 & 23.626 & 47.012 & 70.457 & 94.129 & 88.872 & 77.579 & 69.715 & 64.671 & 72.987 \\
        LDAM	& 69.345 & 95.258 & 99.284 & 96.203 & 0.000 & 27.112 & 47.005 & 71.827 & 94.495 & 87.878 & 79.453 & 65.918 & 64.108 & 72.951 \\
        LADJ	& 70.476 & 95.173 & 99.219 & 96.005 & 0.000 & 26.904 & 53.127 & 73.862 & 94.988 & 89.087 & 80.674 & 67.790 & 67.776 & 71.579 \\
        BalSoft	& 74.027 & 95.381 & 99.022 & 92.467 & 0.000 & 34.631 & 61.714 & 80.696 & 95.618 & 91.967 & 81.893 & 81.108 & 76.353 & 71.505 \\
        Seesaw	& 69.609 & 95.423 & 99.294 & 95.972 & 0.000 & 23.098 & 50.034 & 71.370 & 94.340 & 88.544 & 79.013 & 70.373 & 65.200 & 72.252 \\

        \hline \hline
        \end{tabular}%
    }
\end{center}
\vspace{-1cm}
\end{table}

\begin{table}[t]
\begin{center}
\vspace{-0.8cm}
\caption{Precision (upper part) and recall (lower part) on DALES dataset using RandLA-Net.} \label{tab:precision_recall_dales_randla}
    \resizebox{\textwidth}{!}{%
        \begin{tabular}{|l||c|c|c|c|c|c|c|c|c|}

        \hline
        Method & mean & \textit{ground} & \textit{vegetation} & \textit{cars} & \textit{trucks} & \textit{power lines} & \textit{fences} & \textit{poles} & \textit{buildings} \\
        \hline
        \hline
        \multicolumn{10}{c}{Precision} \\
        \hline
        uni   	& 90.701 & 97.856 & 97.263 & 90.867 & 71.477 & 96.496 & 85.943 & 87.069 & 98.638 \\
        \hline
        invf	& 71.142 & 97.896 & 97.646 & 66.215 & 42.067 & 95.442 & 23.581 & 48.391 & 97.893 \\
        cb	    & 82.464 & 97.798 & 97.622 & 85.886 & 47.852 & 97.718 & 61.025 & 73.094 & 98.719 \\
        invl	& 89.890 & 97.867 & 97.116 & 89.935 & 71.928 & 96.005 & 83.386 & 84.297 & 98.586 \\
        invp	& 89.639 & 98.020 & 97.269 & 90.030 & 65.172 & 97.358 & 83.573 & 87.402 & 98.288 \\
        comf	& 89.957 & 98.173 & 96.700 & 89.207 & 72.176 & 96.306 & 83.013 & 85.582 & 98.497 \\
		\hline	 
        FL	    & 91.748 & 97.877 & 97.184 & 90.854 & 79.932 & 96.151 & 85.964 & 87.477 & 98.542 \\
        LDAM	& 90.526 & 98.068 & 97.285 & 89.086 & 73.628 & 97.549 & 82.176 & 87.901 & 98.519 \\
        LADJ	& 70.000 & 98.275 & 98.143 & 76.875 & 20.733 & 97.030 & 26.483 & 43.926 & 98.533 \\
        BalSoft	& 70.036 & 98.176 & 97.901 & 78.056 & 28.034 & 97.913 & 25.411 & 36.130 & 98.669 \\
        Seesaw	& 80.083 & 98.047 & 97.862 & 84.936 & 40.542 & 96.327 & 50.408 & 74.294 & 98.248 \\

        \hline \hline
        \multicolumn{10}{c}{Recall} \\
        \hline
        uni	    & 81.092 & 99.261 & 95.987 & 90.974 & 45.099 & 94.637 & 58.644 & 66.190 & 97.948 \\
        \hline
        invf	& 90.595 & 98.349 & 91.405 & 95.435 & 59.574 & 95.366 & 94.618 & 92.832 & 97.184 \\
        cb	    & 88.886 & 99.173 & 95.108 & 92.986 & 64.508 & 95.618 & 81.353 & 84.717 & 97.622 \\
        invl	& 81.809 & 99.131 & 95.826 & 92.327 & 46.824 & 93.331 & 60.572 & 68.375 & 98.087 \\
        invp	& 83.744 & 99.052 & 96.020 & 92.092 & 50.205 & 95.442 & 63.392 & 75.519 & 98.229 \\
        comf	& 81.917 & 98.820 & 96.295 & 92.695 & 46.325 & 94.334 & 60.711 & 67.979 & 98.178 \\       
        \hline
        FL	    & 79.831 & 99.246 & 95.952 & 91.141 & 35.843 & 94.351 & 57.136 & 66.944 & 98.034 \\
        LDAM	& 84.187 & 99.078 & 96.114 & 93.815 & 44.914 & 95.434 & 65.467 & 80.438 & 98.239 \\
        LADJ	& 89.902 & 98.719 & 92.729 & 94.625 & 64.755 & 93.314 & 92.231 & 85.257 & 97.584 \\
        BalSoft	& 90.132 & 98.627 & 92.583 & 94.277 & 65.233 & 90.693 & 92.269 & 89.773 & 97.602 \\
        Seesaw	& 86.714 & 99.111 & 94.725 & 92.998 & 55.782 & 94.297 & 81.903 & 76.506 & 98.391 \\
        \hline \hline
        \end{tabular}%
    }
\end{center}
\vspace{-0.5cm}
\end{table}

\begin{table}[b]
\begin{center}
\vspace{-0.5cm}
\caption{Precision (upper part) and recall (lower part) on S3DIS dataset using RandLA-Net.} \label{tab:precision_recall_s3dis_randla}
    \resizebox{\textwidth}{!}{%
        \begin{tabular}{|l||c|c|c|c|c|c|c|c|c|c|c|c|c|c|}
        \hline
        Method & mean & \textit{ceiling} & \textit{floor} & \textit{wall} & \textit{beam} & \textit{column} & \textit{window} & \textit{door} & \textit{chair} & \textit{table} & \textit{bookcase} & \textit{sofa} & \textit{board} & \textit{clutter} \\
        \hline
        \hline
        \multicolumn{15}{c}{Precision} \\
        \hline
        uni	    & 80.728 & 97.141 & 98.861 & 83.381 & 0.000 & 82.797 & 89.759 & 79.471 & 88.871 & 89.566 & 97.199 & 85.960 & 89.194 & 67.263 \\
        \hline
        invf	& 74.180 & 96.600 & 99.173 & 85.167 & 0.000 & 57.375 & 81.855 & 64.004 & 84.179 & 90.240 & 84.962 & 85.450 & 71.991 & 63.341 \\
        cb	    & 76.681 & 95.900 & 99.042 & 84.572 & 0.000 & 70.406 & 82.977 & 75.524 & 88.573 & 86.509 & 86.136 & 86.326 & 75.806 & 65.082 \\
        invl	& 79.798 & 96.674 & 98.868 & 83.692 & 0.000 & 76.523 & 89.892 & 75.811 & 86.614 & 92.015 & 96.899 & 84.885 & 88.008 & 67.499 \\
        invp	& 79.341 & 96.867 & 99.026 & 84.283 & 0.000 & 75.117 & 84.877 & 75.245 & 89.069 & 91.002 & 96.992 & 85.450 & 86.219 & 67.283 \\
        comf	& 79.599 & 96.816 & 98.920 & 84.663 & 0.000 & 71.351 & 93.471 & 71.754 & 89.933 & 92.143 & 94.772 & 86.199 & 89.411 & 65.357 \\       
        \hline
        FL	    & 80.398 & 95.774 & 98.914 & 84.991 & 0.000 & 70.135 & 90.146 & 80.498 & 88.941 & 90.864 & 97.278 & 86.048 & 95.407 & 66.176 \\
        LDAM	& 79.681 & 96.640 & 98.674 & 86.019 & 0.000 & 68.385 & 88.341 & 80.285 & 85.503 & 92.692 & 96.148 & 86.255 & 88.417 & 68.498 \\
        LADJ	& 76.895 & 96.432 & 99.042 & 88.867 & 0.000 & 62.871 & 84.382 & 74.000 & 81.707 & 91.869 & 91.632 & 82.697 & 77.851 & 68.287 \\
        BalSoft	& 75.684 & 96.620 & 98.446 & 88.819 & 0.000 & 53.362 & 83.555 & 74.459 & 84.116 & 91.516 & 92.316 & 81.676 & 72.475 & 66.530 \\
        Seesaw	& 78.258 & 96.112 & 98.849 & 84.981 & 0.000 & 55.685 & 85.652 & 74.141 & 87.210 & 92.814 & 95.411 & 85.078 & 92.204 & 69.220 \\

        \hline \hline
        \multicolumn{15}{c}{Recall} \\
        \hline
        uni	    & 67.684 & 95.740 & 98.126 & 95.696 & 0.000 & 18.042 & 61.720 & 40.713 & 86.710 & 93.928 & 56.758 & 80.057 & 78.186 & 74.220 \\
        \hline
        invf	& 69.340 & 94.399 & 98.188 & 90.999 & 0.000 & 18.021 & 69.679 & 36.903 & 88.979 & 94.459 & 68.539 & 81.233 & 86.815 & 73.206 \\
        cb	    & 70.884 & 95.572 & 97.993 & 93.539 & 0.000 & 29.590 & 69.922 & 37.466 & 87.187 & 93.887 & 86.134 & 79.388 & 81.234 & 69.576 \\
        invl	& 68.100 & 95.381 & 98.452 & 95.501 & 0.000 & 17.143 & 64.180 & 45.263 & 88.244 & 93.442 & 61.481 & 80.720 & 75.135 & 70.359 \\
        invp	& 70.514 & 95.589 & 98.682 & 94.078 & 0.000 & 23.770 & 66.955 & 47.864 & 87.195 & 94.995 & 73.864 & 81.007 & 82.541 & 70.144 \\
        comf	& 68.984 & 95.209 & 98.960 & 94.894 & 0.000 & 26.775 & 62.487 & 38.991 & 85.534 & 94.496 & 63.661 & 80.253 & 80.435 & 75.090 \\
        \hline
        FL	    & 68.694 & 95.433 & 98.397 & 95.012 & 0.000 & 23.507 & 61.979 & 60.422 & 84.257 & 95.105 & 56.009 & 80.752 & 70.954 & 71.191 \\
        LDAM	& 71.517 & 95.208 & 98.180 & 94.015 & 0.000 & 33.208 & 64.448 & 57.590 & 89.033 & 95.321 & 68.197 & 81.362 & 80.974 & 72.185 \\
        LADJ	& 73.371 & 95.328 & 97.573 & 91.665 & 0.000 & 40.847 & 71.001 & 56.323 & 91.549 & 95.437 & 74.703 & 85.956 & 83.689 & 69.754 \\
        BalSoft	& 71.942 & 94.249 & 98.845 & 91.177 & 0.000 & 29.908 & 69.705 & 50.693 & 91.962 & 95.204 & 72.012 & 84.966 & 84.299 & 72.219 \\
        Seesaw	& 69.141 & 95.352 & 98.950 & 94.371 & 0.000 & 24.619 & 66.483 & 54.003 & 88.356 & 92.374 & 58.451 & 81.889 & 74.359 & 69.627 \\
        \hline \hline
        \end{tabular}%
    }
\end{center}
\vspace{-1cm}
\end{table}

\end{document}